\pdfoutput=1

\documentclass[11pt]{article}

\usepackage[final]{acl}

\usepackage{times}
\usepackage{latexsym}

\usepackage[T1]{fontenc}

\usepackage[utf8]{inputenc}

\usepackage{microtype}

\usepackage{inconsolata}

\usepackage{graphicx}

\usepackage{hyperref}
\usepackage{url}
\usepackage{subfig}
\usepackage{xspace}
\usepackage{enumitem}
\usepackage{multirow}
\usepackage{makecell}
\usepackage{wrapfig}
\usepackage{longtable}
\usepackage{booktabs}
\usepackage{amssymb}
\usepackage{amsmath}

\newcommand{\iconleft}{\raisebox{-5pt}{\includegraphics[width=1em]{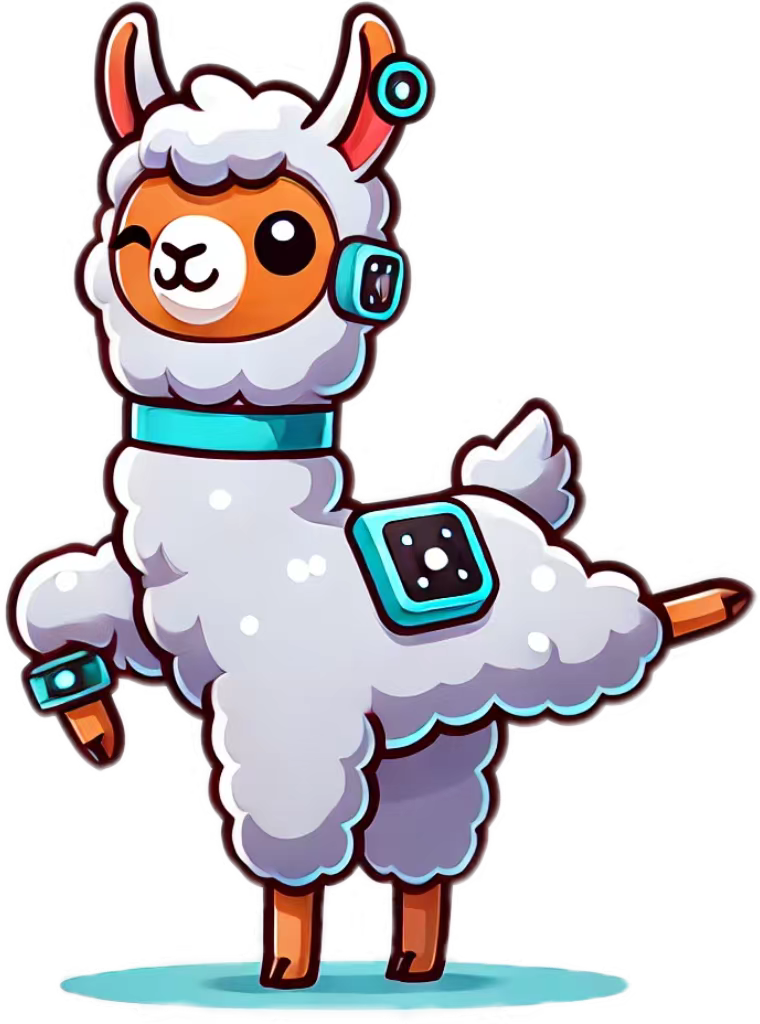}}\thinspace}

\title{\iconleft SensorLLM: Aligning Large Language Models with Motion Sensors for Human Activity Recognition}

\author{%
  Zechen Li$^{1}$,\ \ Shohreh Deldari$^{1}$,\ \ Linyao Chen$^{2}$,\ \ Hao Xue$^{1}$,\ \ Flora D. Salim$^{1}$
   \vspace{1mm} \\
  $^{1}$University of New South Wales, Sydney \\ $^{2}$University of Tokyo  \vspace{1mm} \\
  \texttt{\small \{zechen.li, s.deldari, hao.xue1, flora.salim\}@unsw.edu.au} \\
  \texttt{\small chen-linyao217@g.ecc.u-tokyo.ac.jp}
}

\begin{document}
\maketitle
\begin{abstract}
We introduce SensorLLM, a two-stage framework that enables Large Language Models (LLMs) to perform human activity recognition (HAR) from sensor time-series data. Despite their strong reasoning and generalization capabilities, LLMs remain underutilized for motion sensor data due to the lack of semantic context in time-series, computational constraints, and challenges in processing numerical inputs. SensorLLM addresses these limitations through a Sensor-Language Alignment stage, where the model aligns sensor inputs with trend descriptions. Special tokens are introduced to mark channel boundaries. This alignment enables LLMs to capture numerical variations, channel-specific features, and data of varying durations, without requiring human annotations. In the subsequent Task-Aware Tuning stage, we refine the model for HAR classification, achieving performance that matches or surpasses state-of-the-art methods. Our results demonstrate that SensorLLM evolves into an effective sensor learner, reasoner, and classifier through human-intuitive Sensor-Language Alignment, generalizing across diverse HAR datasets. We believe this work establishes a foundation for future research on time-series and text alignment, paving the way for foundation models in sensor data analysis. Our codes are available at \url{https://github.com/zechenli03/SensorLLM}.
\end{abstract}

\section{Introduction}
\label{introduction}
\begin{figure}[t]
\includegraphics[width=1\columnwidth]{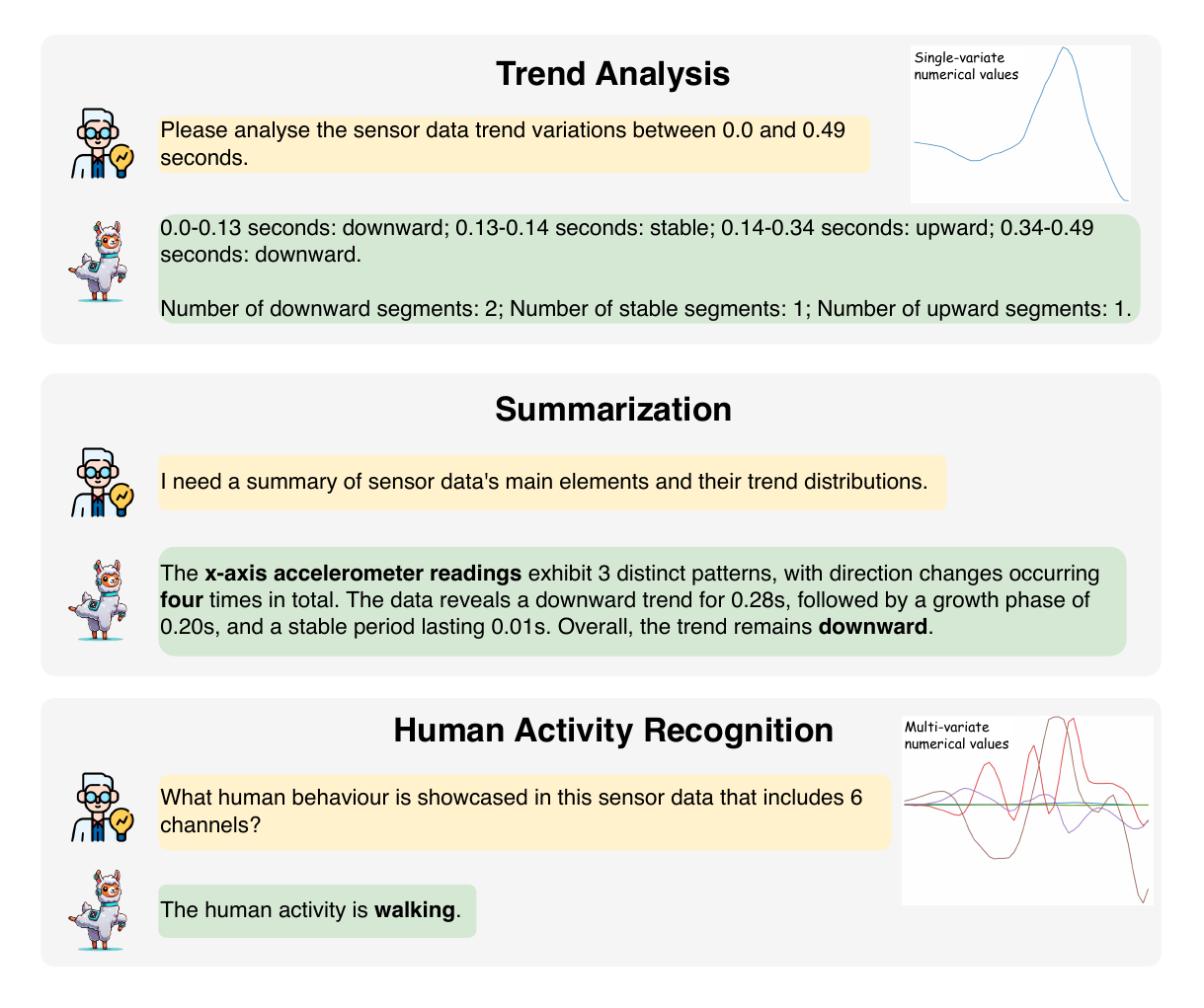}
\vspace{-1em}
\caption{SensorLLM can analyze and summarize trends in captured sensor data, facilitating human activity recognition tasks.}
\label{fig:demo}
\end{figure}

Human Activity Recognition (HAR) is a time-series classification task that maps sensor signals, such as accelerometer and gyroscope data, to human activities. Traditional models like LSTM~\citep{10.1145/3090076, 10.5555/3060832.3060835} and DeepConvLSTM~\citep{s16010115} learn high-level features but are task-specific and struggle to generalize across different sensor configurations and activity sets. In contrast, Large Language Models (LLMs)~\citep{han2021pretrainedmodelspastpresent} have shown remarkable success in integrating diverse data types~\citep{liu2023llava, wu2023nextgpt, yin2023survey}, including text and images.

Enabling LLMs to process wearable sensor data~\citep{jin2023largemodelstimeseries} requires either (1) pretraining or fine-tuning on TS data~\citep{zhou2023fits}, which demands substantial computational resources and is hindered by limited and imbalanced labeled data, or (2) leveraging zero-shot and few-shot prompting by converting sensor data into text~\citep{pmlr-v248-kim24b, ji2024hargptllmszeroshothuman}. The latter approach avoids retraining but introduces key challenges: (i) \textbf{Numerical encoding issues.} Language model tokenizers, designed for text, struggle with numerical values, treating consecutive numbers as independent tokens~\citep{gruver2023llmtime} and failing to preserve temporal dependencies~\citep{spathis2024first}. (ii) \textbf{Sequence length constraints.} Complex numerical sequence often exceeds LLMs’ maximum context length, leading to truncation, information loss, and increased computational costs. (iii) \textbf{Multivariate complexity.} LLMs process univariate inputs, making it difficult to encode multivariate sensor data in a way that retains inter-channel dependencies. (iv) \textbf{Prompt engineering challenges.} Designing effective prompts that enable LLMs to interpret numerical sensor readings, detect trends and classify activities remains a challenge~\citep{liu2023largelanguagemodelsfewshot}.

Unlike image-text pairs, sensor data comprises multi-channel numerical signals with diverse characteristics, making direct interpretation and annotation particularly difficult. Existing methods~\citep{jin2023time, sun2024test} have explored condensed text prototypes for alignment, but these approaches often lack interpretability and require extensive tuning to select suitable prototypes. 

To address these challenges, we propose SensorLLM, a human-intuition inspired framework that aligns wearable sensor data with natural language without modifying the LLM itself. We propose an automatic text generation approach that aligns with human intuition by deriving descriptive trend-based text directly from time-series data using statistical analyses and predefined templates (see Figure~\ref{fig:demo}). This method is precise, scalable, and interpretable, eliminating the need for manual annotations while preserving essential sensor characteristics. SensorLLM follows a two-stage framework:

\paragraph{Sensor-Language Alignment Stage.} We automatically generate question-answer pairs to align sensor time-series with text while preserving temporal features using a pretrained encoder. The resulting embeddings are mapped into a space interpretable by the LLM, mitigating issues associated with text-specific tokenization. Additionally, we introduce \textit{special tokens} for sensor channels, enabling LLMs to effectively capture multi-channel dependencies.

\paragraph{Task-Aware Tuning Stage.} The aligned embeddings are used for HAR, leveraging the LLM’s reasoning capabilities while keeping its parameters frozen. This design extends LLMs beyond their original training, addressing concerns raised by~\citet{tan2024languagemodelsactuallyuseful} regarding their applicability to time-series data. Importantly, our framework naturally supports sensor inputs of varying sequence lengths and arbitrary numbers of channels (i.e., multivariate time-series), a flexibility that prior approaches have struggled to accommodate.

To the best of our knowledge, this is the first approach to integrate sensor data into LLMs for sensor-based analysis and activity recognition. The key contributions of this work are:

\begin{itemize}
    \item We propose a human-intuitive approach for aligning sensor data with descriptive text, eliminating the need for manual annotations. SensorLLM support time-series inputs with varying sequence lengths and channel configurations, allowing broad and realistic HAR evaluation. We evaluate SensorLLM through text similarity metrics, human judgments, and LLM-based assessments, confirming its ability to capture temporal patterns for robust multimodal understanding. 
    
    \item SensorLLM achieves competitive results across five HAR datasets, matching or surpassing state-of-the-art models. Experiments further validate that \textit{modality alignment} and \textit{task-specific prompts} significantly enhance LLM’s ability to interpret and classify sensor data.

    \item We show that SensorLLM maintains strong performance in the Task-Aware Tuning Stage, even when applied to datasets distinct from those used during alignment, highlighting its robustness and generalizability.
\end{itemize}

\section{Related Work}
\label{realted_work}

In this section, we review recent advances in applying LLMs to time-series data, with a focus on three directions: (1) treating general time series as text for LLM-based modeling, (2) leveraging Multimodal Large Language Models (MLLMs) for sensor data, and (3) employing LLMs for Human Activity Recognition (HAR). A broader survey of related work, including deep learning approaches to HAR and additional LLM-based forecasting methods, is provided in Appendix~\ref{apd:related_work}.

\paragraph{LLMs for Time Series as Text.}
While LLMs excel in processing natural language, applying them directly to time-series data poses unique challenges~\citep{spathis2024first}. Certain methods address this by treating time-series signals as raw text, using the same tokenization as natural language. Notable examples include PromptCast~\citep{xue2023promptcast}, which transforms numeric inputs into textual prompts for zero-shot forecasting, and LLMTime~\citep{gruver2024llmtime}, which encodes time-series as numerical strings for GPT-like models. However, due to the lack of specialized tokenizers for numeric sequences, LLMs may fail to capture crucial temporal dependencies and repetitive patterns~\citep{spathis2024first}. To mitigate these issues, several works employ time-series encoders before mapping the resulting embeddings to language model spaces~\citep{liu2024etp, zhou2023tent, xia2024ts2act}, thus aligning sensor embeddings with textual embeddings in a contrastive or supervised manner.

\paragraph{MLLMs for Sensor Data.}
Extending LLMs to non-textual domains has gained traction, particularly through MLLMs that accept inputs beyond text, such as images or speech. For sensor data, the challenge lies in representing continuous signals effectively. \citet{yoon-etal-2024-eyes} propose to ground MLLMs with sensor data via visual prompting. Sensor signals are first visualized as images, guiding the MLLM to analyze the visualized sensor traces alongside task descriptions, which also lower token costs compared to raw-text baselines. Similarly, \citet{moon-etal-2023-imu2clip} introduce IMU2CLIP, which aligns inertial measurement unit streams with text and video in a joint representation space. This approach enables wearable AI applications like motion-based media search and LM-based multimodal reasoning, showcasing how sensor data can be integrated into broader multimodal frameworks. 

\paragraph{LLMs for Human Activity Recognition.} Despite the success of LLMs in NLP, applying them to HAR remains challenging, as inertial signals are difficult to ground in language. \citet{xia2023unsupervised} introduce an unsupervised pipeline that employs a two-stage prompting scheme to infer activities from object sequences without manual descriptions. IMUGPT~\cite{Leng2023generating} follows a data-synthesis approach: it prompts LLMs to generate diverse textual activity descriptions, converts them into 3D motion sequences, and then renders virtual IMU streams for training. HARGPT~\cite{ji2024hargptllmszeroshothuman} directly prompts LLMs to classify sensor time-series data, downsampling signals to 10Hz to enable zero-shot HAR from raw IMU inputs, and demonstrates that on simple tasks LLMs can match or even surpass traditional and deep learning models without domain-specific adaptation. ZARA~\cite{li2025zarazeroshotmotiontimeseries} instead frames HAR as an agentic LLM workflow, supporting plug-and-play zero-shot motion time-series analysis with knowledge-guided reasoning and retrieval-enhanced evidence.
\section{Methods}
\label{methods}

\begin{figure*}[t]
\begin{center}
 \includegraphics[width=1\textwidth]{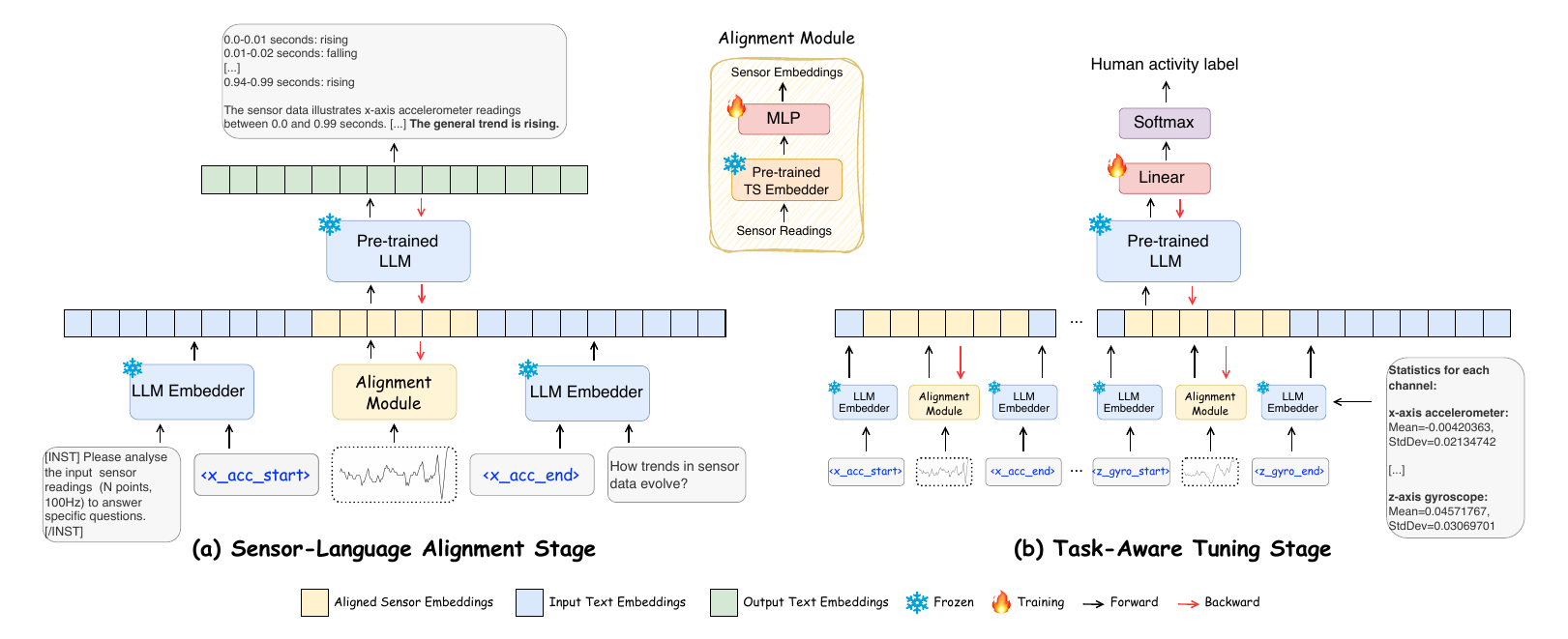}
\end{center}
\vspace{-0.5em}
\caption{Our proposed SensorLLM framework: \textbf{(a) Sensor-Language Alignment Stage}, where a generative model aligns sensor readings with automatically generated text; \textbf{(b) Task-Aware Tuning Stage}, where a classification model leverages the aligned modalities to perform HAR.}
\label{fig:sensorllm}
\end{figure*}

In this work, we propose SensorLLM, a two-stage framework that aligns wearable sensor data with descriptive text by high-precision question-answering pairs created without any human annotations and tailored for wearable sensor reasoning. Our aim is to build a multimodal model capable of interpreting and reasoning over time-series (TS) signals. As shown in Figure \ref{fig:sensorllm}, SensorLLM consists of three core components: (1) a pretrained LLM, (2) a pretrained TS embedder, and (3) a lightweight MLP alignment module.

In the Sensor-Language Alignment Stage, a generative model aligns sensor readings with text , and in the Task–Aware Tuning stage, a lightweight classifier is added on top of the LLM to perform HAR. Crucially, only the alignment MLP and this classifier are trainable, while the backbone LLM and TS embedder remain frozen. This design results in just 5.67\% (535.9 M) of parameters being fine-tuned in the first stage and 0.12\% (10.5 M) in the second, making training extremely efficient.

\subsection{Sensor-Text Data Generation}
\label{Sensor-Text}
Aligning time-series data with text for LLM-based tasks is challenging due to the lack of rich semantic labels beyond class annotations, making manual annotation impractical~\citep{deldari2024crossl, haresamudram2024limitations}. While prior works rely on predefined text prototypes~\citep{sun2024test, jin2023time}, we aim for a more human-intuitive representation of sensor data.

We argue that time-series data inherently contains semantic patterns that can be expressed through descriptive text, from simple numerical trends to statistical insights. To achieve this, we automatically generate descriptive text by analyzing observed trends and fluctuations in the data. Using predefined templates, we construct diverse question-answer (QA) pairs that capture trend changes while ensuring accuracy and scalability. These templates (see Appendix~\ref{apd:data_generation}) are randomly combined to enhance diversity. For example:
\begin{itemize}
    \item[(1)] The time-series data represents readings taken from a \texttt{<S>} sensor between \texttt{<$t_{s}$>} and \texttt{<$t_{e}$>} seconds.
    \item[(2)] To sum up, the data exhibited a \texttt{<T>} trend for a cumulative period of \texttt{<$t_{t}$>} seconds.
\end{itemize}

\noindent where \texttt{T} and \texttt{S} denote specific trends and sensor types, and \texttt{t} corresponds to numerical values.

\subsection{Sensor-Language Alignment}
\label{stage1}
As shown in Figure~\ref{fig:sensorllm} (a), the Sensor-Language Alignment stage employs a generative model to create multimodal sentences that combine single-channel sensor readings with textual descriptions. The sensor data is represented as a matrix $\mathbf{X} \in \mathbb{R}^{C \times T}$, where $C$ is the number of channels and $T$ is the sequence length. Each channel’s data, denoted as $\mathbf{X}^{c}$, is processed independently to retain channel-specific characteristics. The data is segmented into non-overlapping segments $\mathbf{X}_{S}^{c}$, where $S$ is the total number of segments. Each segment $x_s$ is assigned a random length $l$ within a predefined range, encouraging the model to learn from both short-term fluctuations and long-term trends.

We use Chronos~\citep{ansari2024chronos} as the TS encoder to generate segment embeddings $\Hat{x}_{s} \in \mathbb{R}^{(l+1) \times d{ts}}$, where $d_{ts}$ is the embedding dimension and $(l+1)$ accounts for the \texttt{[EOS]} token added during Chronos tokenization (Appendix~\ref{apd:chronos}). Before feeding segments into Chronos, we apply instance normalization: $\tilde{x}_s = \frac{x_s - \text{mean}(x_s)}{\text{std}(x_s)}$. For the language backbone, we use LLaMA3-8B~\citep{touvron2023llamaopenefficientfoundation}.

\paragraph{Alignment Module.} 
To transform TS embeddings $\Hat{x}_{s}$ into text-aligned embeddings $\Hat{a}_{s} \in \mathbb{R}^{(l+1) \times D}$ for downstream tasks, we introduce an alignment projection module. This module, implemented as a multi-layer perceptron (MLP), first maps sensor embeddings to an intermediate space of dimension $d_m$ and then projects them to the target dimension $D$. Formally,  

\begin{equation}
\Hat{a}_{s} = \mathbf{W}_2 \cdot \sigma (\mathbf{W}_1 \Hat{x}_{s} + \mathbf{b}_1) + \mathbf{b}_2,
\end{equation}

\noindent where $\mathbf{W}_1 \in \mathbb{R}^{d_m \times d_{ts}}$ and $\mathbf{W}_2 \in \mathbb{R}^{D \times d_m}$ are learnable weights, $\mathbf{b}_1$ and $\mathbf{b}_2$ are biases, and $\sigma$ is the GELU activation function~\citep{hendrycks2016gelu}. This projection ensures that the transformed embeddings $\Hat{a}_{s}$ are semantically aligned with the text embedding space, making them suitable for tasks such as text generation and classification.

\paragraph{Input Embedding.}
To integrate sensor data into the LLM, we introduce two special tokens per sensor channel (e.g., \texttt{<x\_acc\_start>} and \texttt{<x\_acc\_end>} for the x-axis accelerometer), extending the LLM’s embedding matrix from $\mathbf{E} \in \mathbb{R}^{V \times D}$ to $\mathbf{E} \in \mathbb{R}^{V^\prime \times D}$, where $V^\prime = V + 2c$, with $V$ as the vocabulary size and $c$ as the number of channels. These special token embeddings are concatenated with the aligned sensor embeddings. The final combined sensor representation $\Hat{o}_{s} \in \mathbb{R}^{(l+3) \times D}$ is then concatenated with instruction and question embeddings to form the full input sequence $\Hat{z} \in \mathbb{R}^{k \times D}$, where $k$ is the total number of tokens.

\paragraph{Loss Function.}
SensorLLM processes an input sequence $\mathbf{Z}_s = \{z_s^i\}_{i=1}^K$ consisting of sensor and text embeddings and generates an output sequence $\mathbf{Z}_t = \{z_t^i\}_{i=1}^N$, where $z_s^i, z_t^i \in V^\prime$, and $K$ and $N$ represent the number of input and output tokens, respectively. The model is trained using a causal language modeling objective, predicting the next token based on previous ones. The optimization minimizes the negative log-likelihood:  

\begin{equation}
\mathcal{L}_{gen} = -\sum_{i=0}^{N-1} \log P(z_t^i | Z_t^{<i}, z_s).
\end{equation}

\noindent Loss is computed only on generated tokens, ensuring SensorLLM effectively integrates sensor and text embeddings to produce coherent, contextually appropriate responses.

\subsection{Task-Aware Tuning}
\label{stage2}

As shown in Figure~\ref{fig:sensorllm} (b), the Task-Aware Tuning stage refines the multimodal sensor-text embeddings for HAR. This stage integrates multi-channel sensor readings with activity labels, aligning temporal patterns with human activities. The input sensor data $\mathbf{X}$ is segmented into overlapping windows of size $L$ with a 50\% overlap~\citep{10.1145/3267242.3267246}, forming segments $\mathbf{X}_S \in \mathbb{R}^{S \times C \times L}$, where $S$ is the number of segments and $C$ is the number of channels. The pretrained alignment module from the first stage maps sensor data to activity labels, preserving inter-channel dependencies while learning activity-related patterns.

\paragraph{Input Embedding.}
For each sensor channel $c$, we retrieve its aligned sensor embeddings $\Hat{o}_{s}^c$ from the pretrained alignment module. These embeddings are then concatenated across all channels, along with their corresponding statistical features (mean and variance), to form the final input embedding:  

\begin{equation}
\Hat{z} = \Hat{o}_{s}^1 \oplus \Hat{o}_{s}^2 \oplus \dots \oplus \Hat{o}_{s}^C \oplus \Hat{z}_{\text{stat}},
\end{equation}

\noindent where $\Hat{z}_{\text{stat}}$ represents the statistical information, and $C$ is the number of channels. This ensures the model integrates both temporal and statistical characteristics for HAR.

\paragraph{Loss Function.}
The input token sequence is processed by the LLM, yielding a latent representation $\mathbf{H} \in \mathbb{R}^{K \times D}$, where $K$ is the number of tokens and $D$ is the embedding dimension. Due to causal masking, we extract the final hidden state, $\mathbf{h} = \mathbf{H}_K$, which encodes all preceding token information. This pooled vector is passed through a fully connected layer to produce a prediction vector of size $M$, where $M$ is the number of activity classes. The final class probabilities $\hat{y}_i$ are obtained via the softmax function, and the model is optimized using cross-entropy loss:

\begin{equation}
\mathcal{L}_{cls} = -\sum_{i=0}^{M-1}y_i\log \hat{y}_i,
\end{equation}

\noindent where $y_i$ is the ground truth label.

\begin{table*}[t]
    \centering
    \resizebox{\textwidth}{!}{
    \begin{tabular}{l|cccccccccc}
        \toprule
        \multirow{3}{*}{\textbf{Metric}} & \multicolumn{2}{c}{\textbf{USC-HAD}} & \multicolumn{2}{c}{\textbf{UCI-HAR}}  & \multicolumn{2}{c}{\textbf{PAMAP2}}  & \multicolumn{2}{c}{\textbf{MHealth}} & \multicolumn{2}{c}{\textbf{CAPTURE-24}}\\
        \cmidrule(lr){2-3} \cmidrule(lr){4-5} \cmidrule(lr){6-7} \cmidrule(lr){8-9} \cmidrule(lr){10-11}
        & \textbf{GPT-4o} & \textbf{Ours} & \textbf{GPT-4o} & \textbf{Ours} & \textbf{GPT-4o} & \textbf{Ours} & \textbf{GPT-4o} & \textbf{Ours} & \textbf{GPT-4o} & \textbf{Ours}  \\
        \midrule
        BLEU-1  &41.43  &\textbf{57.68}    &37.97       &\textbf{56.78}   &46.35  &\textbf{60.20} &49.97   & \textbf{61.38}  &46.58  &\textbf{57.10}  \\
        ROUGE-1 &54.92   & \textbf{68.32}   & 51.24  & \textbf{67.63}  &58.08 &\textbf{69.92} &61.11   & \textbf{71.20}  &58.21  &\textbf{68.11}    \\
        ROUGE-L &49.00  & \textbf{64.17}  & 44.88  &\textbf{63.05}  &50.30  &\textbf{66.25}  &51.99   & \textbf{67.83}  &48.88  & \textbf{60.90}   \\
        METEOR  &30.51  & \textbf{45.95} & 26.93  & \textbf{45.81}     &37.17 &\textbf{52.21}  &38.50   & \textbf{51.73}  &31.16  & \textbf{40.51}   \\
        SBERT  &77.22   &  \textbf{86.09}  & 76.05  & \textbf{85.01}  &82.71 &\textbf{87.31}  &83.15   &\textbf{86.66}  &83.11  &  \textbf{84.83}   \\
        SimCSE &86.96 &\textbf{93.09} &90.23 &\textbf{92.51} & 89.64 &\textbf{93.82}  &92.10   &\textbf{93.38 }  &90.10  &  \textbf{92.20}  \\  
        \midrule
        GPT-4o  &1.67   &\textbf{3.11}    &1.61   &\textbf{3.20}   &1.90  &\textbf{3.77}   &1.69   &\textbf{3.69}  &1.70  &    \textbf{2.32}   \\
        Human  &2.10  &\textbf{4.16}    &1.94   &\textbf{4.04}   &2.38    &\textbf{4.70}   & 1.74  &\textbf{4.56}   &2.30  &  \textbf{3.10}    \\
        \bottomrule
    \end{tabular}
    }
    \caption{Evaluation of Sensor Data Understanding tasks. The column \textit{GPT-4o} denotes trend descriptions generated by GPT-4o, while the row \textit{GPT-4o} indicates evaluations conducted by GPT-4o on the model outputs.}
    \label{tab:stage1_results}
\end{table*}

\section{Experiments}
\label{experiments}
In this section, we evaluate SensorLLM in enabling LLMs to interpret, reason about, and classify sensor data for HAR tasks. All experiments are conducted on NVIDIA A100-80G GPUs. To assess the LLM’s ability to learn and generalize from raw sensor inputs, we ensure that the same training and testing subjects are used in both the Sensor-Language Alignment and Task-Aware Tuning stages. This guarantees that test data in the second stage remains unseen during alignment, ensuring a fair evaluation of generalization. We select Chronos as the TS embedder because it has not been pre-trained on motion sensor data, making it an ideal candidate for evaluating the robustness of our approach in adapting to raw, domain-agnostic sensor signals.

\subsection{Datasets}
\label{datasets}
To evaluate the effectiveness and generalizability of SensorLLM, we conduct experiments on five publicly available HAR datasets: USC-HAD~\citep{10.1145/2370216.2370438}, UCI-HAR~\citep{Anguita2013APD}, PAMAP2~\citep{6246152}, MHealth~\citep{Baos2014mHealthDroidAN}, and CAPTURE-24~\citep{chan2024capture24largedatasetwristworn}. These datasets vary widely in subject counts, sensor placement, sampling rates, channel configurations, and activity types, covering both controlled laboratory conditions and free-living environments. All datasets are publicly available, containing no personally identifiable information, thus posing minimal ethical or privacy concerns.

We use subject-independent splits for all datasets except UCI-HAR, which comes with a fixed split. In all other datasets, training and test sets come from different subjects, ensuring the model is evaluated on unseen users. Full dataset details, including subject count, sensor configurations, data splits, activity classes, preprocessing steps, and windowing strategies, are provided in Appendix~\ref{apd:dataset}.

\subsection{Sensor Data Understanding}
\label{trend_analysis}

\paragraph{Setup.}
All datasets are trained using the same parameters in the Sensor-Language Alignment Stage: a learning rate of 2e-3, 8 epochs, batch size of 4, gradient accumulation steps of 8, and a maximum sequence length of 8192 for CAPTURE-24 and 4096 for others.

\paragraph{Evaluation Metrics.}
We assess the performance of SensorLLM in the sensor–language alignment stage by comparing its ability to generate trend descriptions from sensor data with that of the advanced GPT-4o~\footnote{gpt-4o-2024-08-06~\citep{openai2024gpt4technicalreport}}.
GPT-4o generates responses using a predefined prompt (Appendix~\ref{apd:gpt_gen}). We adopt three evaluation methods:

\begin{itemize}
    \item \textbf{NLP Metrics.} We use BLEU-1~\citep{10.3115/1073083.1073135}, ROUGE-1, ROUGE-L~\citep{lin-2004-rouge}, and METEOR~\citep{banerjee-lavie-2005-meteor} to measure surface-level similarity and n-gram overlap. For deeper semantic alignment and factual correctness, we adopt SBERT~\citep{Reimers2019SentenceBERTSE} and SimCSE~\citep{gao2021simcse}.
    
    \item \textbf{GPT-4o Evaluation.} GPT-4o rates the generated trend descriptions on a scale of 1 to 5 (with 5 being the highest) by comparing each output to ground truth and providing explanatory feedback. As an advanced LLM, its evaluation ensures a semantic assessment of trend comprehension.
    
    \item \textbf{Human Evaluation.} Five time-series experts (PhD students, postdocs, and academics) score accuracy and quality using the same criteria as GPT-4o, providing a human-centered perspective on the model's outputs.
\end{itemize}

Appendix~\ref{apd:metrics} details all metrics and scoring criteria. We randomly sample 200 instances per dataset for both SensorLLM and GPT-4o, then average the results for comparison. Because reading and comparing lengthy sequences is difficult for human annotators, we conduct human evaluation on 20 shorter sequences per dataset (each containing at most 50 time steps).

\paragraph{Results.} Table~\ref{tab:stage1_results} compares SensorLLM and GPT-4o on the Sensor Data Trend Analysis tasks, showing that our model consistently outperforms GPT-4o across all metrics. BLEU-1, ROUGE-1, ROUGE-L, and METEOR primarily focus on surface-level lexical or n-gram overlaps. SBERT and SimCSE can capture factual correctness or deeper semantic similarities. Across all metrics, SensorLLM generates trend descriptions more closely aligned with the ground truth. GPT-4o evaluations further highlight SensorLLM’s superior ability to capture trend details and coherence, whereas GPT-4o struggles with complex numerical data and trend observations~\citep{yehudai2024transformerscountn}. Human evaluation also favors SensorLLM, particularly for shorter sequences. CAPTURE-24 results are weaker compared to other datasets, likely due to its longer sequences being trained with the same parameters. Overall, these findings validate the effectiveness of our Sensor-Language Alignment method in enhancing LLMs’ ability to interpret complex numerical sequence. Appendix~\ref{apd:stage1_results} provides qualitative examples of outputs from both models.

\begin{table*}[t]
    \centering
    \resizebox{\textwidth}{!}{
    \begin{tabular}{l|cccccccccc}
        \toprule
        \multirow{3}{*}{\textbf{Method}}
          &\multicolumn{2}{c}{\textbf{USC-HAD}}    & \multicolumn{2}{c}{\textbf{UCI-HAR}}   & \multicolumn{2}{c}{\textbf{PAMAP2}}    & \multicolumn{2}{c}{\textbf{MHealth}}    &\multicolumn{2}{c}{\textbf{CAPTURE-24}} \\
        \cmidrule(lr){2-3} \cmidrule(lr){4-5} \cmidrule(lr){6-7} \cmidrule(lr){8-9} \cmidrule(lr){10-11}
          & \textbf{F1-macro} & \textbf{Accuracy}  & \textbf{F1-macro} & \textbf{Accuracy}  & \textbf{F1-macro} & \textbf{Accuracy} & \textbf{F1-macro} & \textbf{Accuracy} & \textbf{F1-macro} & \textbf{Accuracy} \\
        \midrule
        PatchTST  & 45.2\textsubscript{\(\pm 1.48\)}    &    45.6\textsubscript{\(\pm 2.19\)}          & 86.8\textsubscript{\(\pm 0.84\)}    &   86.0\textsubscript{\(\pm 0.71\)}             & 82.0\textsubscript{\(\pm 0.71\)}    &    81.2\textsubscript{\(\pm 0.84\)}               &  80.0\textsubscript{\(\pm 1.58\)}    &   79.4\textsubscript{\(\pm 1.34\)}             &  35.6\textsubscript{\(\pm 0.89\)}    & 66.2\textsubscript{\(\pm 1.10\)}               \\
        Ns-Transformer  & 52.6\textsubscript{\(\pm 2.30\)}    &   51.8\textsubscript{\(\pm 2.86\)}      & 88.0\textsubscript{\(\pm 0.71\)}    &87.4\textsubscript{\(\pm 0.55\)}                 & 78.8\textsubscript{\(\pm 0.84\)}    &   78.8\textsubscript{\(\pm 0.84\)}                & 77.2\textsubscript{\(\pm 1.48\)}     &   75.8\textsubscript{\(\pm 1.48\)}            & 34.8\textsubscript{\(\pm 1.10\)}    &   65.4\textsubscript{\(\pm 0.55\)}            \\
        Informer   &51.2\textsubscript{\(\pm 1.30\)}      & 51.6\textsubscript{\(\pm 1.52\)}              &   86.6\textsubscript{\(\pm 1.14\)}     &   86.4\textsubscript{\(\pm 0.89\)}              & 78.0\textsubscript{\(\pm 1.58\)}     &  78.6\textsubscript{\(\pm 1.34\)}             & 74.0\textsubscript{\(\pm 0.71\)}     &  72.8\textsubscript{\(\pm 0.84\)}             &35.6\textsubscript{\(\pm 0.55\)}     &66.8\textsubscript{\(\pm 0.84\)}             \\
        Transformer & 49.6\textsubscript{\(\pm 1.67\)}     &  50.6\textsubscript{\(\pm 0.55\)}             &   85.4\textsubscript{\(\pm 0.89\)}     &   85.2\textsubscript{\(\pm 1.10\)}             &  77.0\textsubscript{\(\pm 0.71\)}     &    77.6\textsubscript{\(\pm 0.89\)}            & 75.2\textsubscript{\(\pm 1.30\)}    &  74.6\textsubscript{\(\pm 1.34\)}             & 32.8\textsubscript{\(\pm 0.84\)}    & 65.4\textsubscript{\(\pm 0.89\)}             \\
        iTransformer    & 48.4\textsubscript{\(\pm 1.82\)}     &  49.6\textsubscript{\(\pm 1.67\)}             &  81.8\textsubscript{\(\pm 0.84\)}     & 81.8\textsubscript{\(\pm 0.84\)}                & 76.6\textsubscript{\(\pm 0.55\)}      &  75.8\textsubscript{\(\pm 0.45\)}             & 80.4\textsubscript{\(\pm 1.14\)}     &    80.0\textsubscript{\(\pm 1.22\)}          & 19.8\textsubscript{\(\pm 0.84\)}    &   62.4\textsubscript{\(\pm 0.89\)}           \\
        TimesNet   & 52.2\textsubscript{\(\pm 2.39\)}     &    52.6\textsubscript{\(\pm 2.07\)}            &  87.4\textsubscript{\(\pm 1.14\)}     &   86.6\textsubscript{\(\pm 1.14\)}               & 76.2\textsubscript{\(\pm 1.92\)}     & 77.4\textsubscript{\(\pm 1.14\)}               &  78.4\textsubscript{\(\pm 1.52\)}     &    77.2\textsubscript{\(\pm 1.48\)}           & 34.8\textsubscript{\(\pm 0.84\)}    & 65.8\textsubscript{\(\pm 1.79\)}              \\
        GPT4TS  & 54.2\textsubscript{\(\pm 2.05\)}     &  56.0\textsubscript{\(\pm 1.58\)}             &  88.2\textsubscript{\(\pm 0.84\)}     &    87.6\textsubscript{\(\pm 0.55\)}            & 80.4\textsubscript{\(\pm 0.89\)}      &   79.8\textsubscript{\(\pm 0.45\)}           &  76.4\textsubscript{\(\pm 1.14\)}     &   75.4\textsubscript{\(\pm 1.14\)}          & 32.8\textsubscript{\(\pm 1.10\)}    &62.2\textsubscript{\(\pm 1.92\)}             \\
        Chronos+MLP & 44.2\textsubscript{\(\pm 1.30\)}      &    44.0\textsubscript{\(\pm 0.71\)}          &  82.2\textsubscript{\(\pm 0.84\)}     &      81.2\textsubscript{\(\pm 0.84\)}           & 79.8\textsubscript{\(\pm 0.45\)}     &   79.8\textsubscript{\(\pm 0.45\)}            & 83.0\textsubscript{\(\pm 0.71\)}     &   82.0\textsubscript{\(\pm 0.71\)}            & 38.0\textsubscript{\(\pm 0.71\)}    &     68.2\textsubscript{\(\pm 0.84\)}         \\
        DeepConvLSTM    & 48.8\textsubscript{\(\pm 2.39\)}     &    50.6\textsubscript{\(\pm 2.41\)}          &  89.2\textsubscript{\(\pm 0.84\)}      &    89.2\textsubscript{\(\pm 0.84\)}             &  78.4\textsubscript{\(\pm 1.52\)}      &   78.2\textsubscript{\(\pm 1.10\)}           & 75.0\textsubscript{\(\pm 1.87\)}     &    76.0\textsubscript{\(\pm 1.00\)}          & 40.4\textsubscript{\(\pm 0.89\)}    &    69.4\textsubscript{\(\pm 1.14\)}           \\
        DeepConvLSTMAtt & 54.0\textsubscript{\(\pm 2.12\)}     &    54.4\textsubscript{\(\pm 3.21\)}           &  89.6\textsubscript{\(\pm 1.14\)}     &     89.4\textsubscript{\(\pm 1.14\)}          & 79.2\textsubscript{\(\pm 1.30\)}     &    79.6\textsubscript{\(\pm 1.14\)}            & 77.4\textsubscript{\(\pm 2.19\)}     &    76.8\textsubscript{\(\pm 1.48\)}         & 41.4\textsubscript{\(\pm 0.55\)}     &      70.4\textsubscript{\(\pm 0.55\)}       \\
        Attend &  \underline{60.2}\textsubscript{\(\pm 2.17\)}      &\underline{60.8}\textsubscript{\(\pm 1.92\)}             &  \textbf{93.2}\textsubscript{\(\pm 0.84\)}       &      \textbf{92.8}\textsubscript{\(\pm 0.45\)}          & \underline{84.6}\textsubscript{\(\pm 1.14\)}     &  \underline{85.0}\textsubscript{\(\pm 0.71\)}              & \underline{83.4}\textsubscript{\(\pm 1.14\)}     &   \underline{82.6}\textsubscript{\(\pm 1.14\)}           & \underline{43.6}\textsubscript{\(\pm 0.55\)}     &   \underline{71.0}\textsubscript{\(\pm 0.71\)}            \\
        \midrule
        SensorLLM   & \textbf{61.2}\textsubscript{\(\pm 3.56\)}      &    \textbf{62.6}\textsubscript{\(\pm 3.36\)}            &  \underline{91.2}\textsubscript{\(\pm 1.48\)}      &    \underline{90.8}\textsubscript{\(\pm 1.30\)}           & \textbf{86.2}\textsubscript{\(\pm 1.48\)}      &   \textbf{87.2}\textsubscript{\(\pm 0.84\)}           & \textbf{89.4}\textsubscript{\(\pm 3.85\)}    & \textbf{89.0}\textsubscript{\(\pm 3.54\)}             & \textbf{48.6}\textsubscript{\(\pm 1.14\)}     & \textbf{72.0}\textsubscript{\(\pm 0.71\)}           \\
        \bottomrule
    \end{tabular}
    }
    \caption{F1-macro and accuracy scores (\%) for the Human Activity Recognition tasks, presented as the mean and standard deviation over 5 random repetitions. \textbf{Bold} for the best and \underline{underline} for the second-best.}
    \label{tab:stage2_results}
\end{table*}

\subsection{Human Activity Recognition}
\label{har}

\paragraph{Setup.}
In this section, we evaluate the performance of SensorLLM on HAR tasks. Each experiment is run for five trials using 8 training epochs, a batch size of 4, gradient accumulation steps of 8, and a maximum sequence length of 4096. We report both F1-macro (Appendix~\ref{apd:metrics_stage2}) and accuracy to account for class imbalance and overall prediction performance across different activity categories.

\paragraph{Baselines.}
We benchmark SensorLLM against 11 baselines across two categories:  
(i) \textit{TS models}—Transformer~\citep{NIPS2017_3f5ee243}, Informer~\citep{haoyietal-informer-2021}, NS-Transformer~\citep{liu2022non}, PatchTST~\citep{Yuqietal-2023-PatchTST}, TimesNet~\citep{wu2023timesnet}, and iTransformer~\citep{liu2024itransformer};  
(ii) \textit{HAR models}—DeepConvLSTM~\citep{s16010115}, DeepConvLSTMAttn~\citep{10.1145/3267242.3267287}, and Attend~\citep{10.1145/3448083}.  
We also include Chronos+MLP and GPT4TS~\citep{zhou2023fits} for a more comprehensive comparison. Full baseline details are in Appendix~\ref{apd:baselines}.

\paragraph{Results.}
Table~\ref{tab:stage2_results} reports the F1-macro and accuracy scores (\%) averaged over five random runs. SensorLLM achieves the best performance on four out of five datasets (USC-HAD, PAMAP2, MHealth, CAPTURE-24), and ranks second on UCI-HAR, indicating strong performance across diverse sensor settings. On CAPTURE-24, SensorLLM reaches 48.6\% F1-macro, a +5.0\% absolute gain over Attend (43.6\%). It also leads on USC-HAD (61.2\%, +1.0\%), PAMAP2 (86.2\%, +1.6\%), and MHealth (89.4\%, +6.0\%), establishing state-of-the-art performance on these datasets. On UCI-HAR, SensorLLM achieves 91.2\%, slightly below Attend (93.2\%).

However, Chronos+MLP offers only a marginal improvement over iTransformer on UCI-HAR (82.2\% vs. 81.8\%), suggesting that Chronos embeddings alone are of limited utility for this dataset. In contrast, using the same time-series encoder (Chronos), SensorLLM substantially boosts both F1-macro and accuracy, with especially large gains on challenging, real-world, long-sequence settings such as CAPTURE-24 and MHealth, highlighting the effectiveness of our alignment strategy in enabling LLMs to understand and classify sensor data. Overall, the consistent gains in both F1-macro and accuracy indicate balanced per-class performance alongside strong overall prediction, with robust generalization across sensor configurations, activity types, and data-collection environments.

\section{Ablation Studies}
\label{ablation}

\begin{table}[t]
\centering
\resizebox{\linewidth}{!}{
  \begin{tabular}{l|cccc}
		\toprule
        \multirow{3}{*}{\textbf{Dataset}}
        & \multicolumn{2}{c}{\textbf{Task-only}} & \multicolumn{2}{c}{\textbf{SensorLLM}} \\
        \cmidrule(lr){2-3} \cmidrule(lr){4-5} 
        & \textbf{w/o prompts} & \textbf{w/ prompts} & \textbf{w/o prompts} & \textbf{w/ prompts} \\
        \midrule
        USC-HAD    & 43.4\textsubscript{\(\pm 2.88\)} & \textbf{45.0}\textsubscript{\(\pm 1.58\)}     & 49.6\textsubscript{\(\pm 1.67\)}      & \textbf{61.2}\textsubscript{\(\pm 3.56\)}   \\  
        UCI-HAR    & 80.0\textsubscript{\(\pm 2.12\)} & \textbf{82.0}\textsubscript{\(\pm 1.58\)}     & 89.2\textsubscript{\(\pm 1.10\)}      & \textbf{91.2}\textsubscript{\(\pm 1.48\)}   \\  
        PAMAP2   & 74.2\textsubscript{\(\pm 2.28\)} & \textbf{75.4}\textsubscript{\(\pm 3.05\)}    & 83.0\textsubscript{\(\pm 0.71\)}      & \textbf{86.2}\textsubscript{\(\pm 1.48\)}   \\  
        MHealth   & 76.6\textsubscript{\(\pm 1.34\)} & \textbf{77.4}\textsubscript{\(\pm 3.13\)}     & 86.6\textsubscript{\(\pm 1.14\)}      & \textbf{89.4}\textsubscript{\(\pm 3.85\)}   \\  
        CAPTURE-24 & 44.8\textsubscript{\(\pm 0.84\)} & \textbf{46.0}\textsubscript{\(\pm 0.71\)}     & 47.2\textsubscript{\(\pm 0.84\)}      & \textbf{48.6}\textsubscript{\(\pm 1.14\)}  \\  
        \bottomrule
	\end{tabular}
}
\caption{F1-macro scores for models trained with and without text prompts. \textit{Task-only} refers to conducting Task-Aware Tuning directly bypassing the alignment stage.}
\label{tab:prompt_results}
\end{table}

\paragraph{Removing Alignment Hurts.}  
To assess the role of sensor–language alignment, we include the Chronos+MLP baseline (Section~\ref{har}) to demonstrate that SensorLLM’s performance is not solely due to the strength of the Chronos encoder. We further compare SensorLLM with a Task-only variant that skips the Sensor–Language Alignment stage and directly feeds Chronos embeddings into the LLM for HAR. As shown in Table~\ref{tab:prompt_results}, SensorLLM consistently outperforms the Task-only model across all five datasets, regardless of whether textual prompts are included. Notably, the Task-only model often performs comparably to or worse than traditional TS baselines, underscoring the critical role of alignment. These results confirm that Chronos embeddings alone are insufficient for optimal HAR performance, and that our alignment stage is essential for enabling the LLM to effectively interpret sensor data.

\paragraph{Textual Prompts Enhance HAR.}  
To assess the role of additional textual information (e.g., statistical features for each sensor channel) in the Task-Aware Tuning Stage, we compared SensorLLM’s performance with and without prompts. As shown in Table~\ref{tab:prompt_results}, incorporating prompts consistently improves F1-macro scores across all datasets, with a more pronounced effect in the full SensorLLM architecture. This demonstrates that the model effectively integrates sensor and textual data, enhancing its ability to capture complex temporal patterns. The results highlight the benefits of multimodal inputs, which enrich sensor data representations and improve HAR accuracy. More broadly, the ability to process both sensor signals and textual prompts not only enhances classification performance but also underscores the potential of LLMs for more generalizable and interpretable sensor-driven applications.

\begin{figure}[t]
    \centering
    \includegraphics[width=\linewidth]{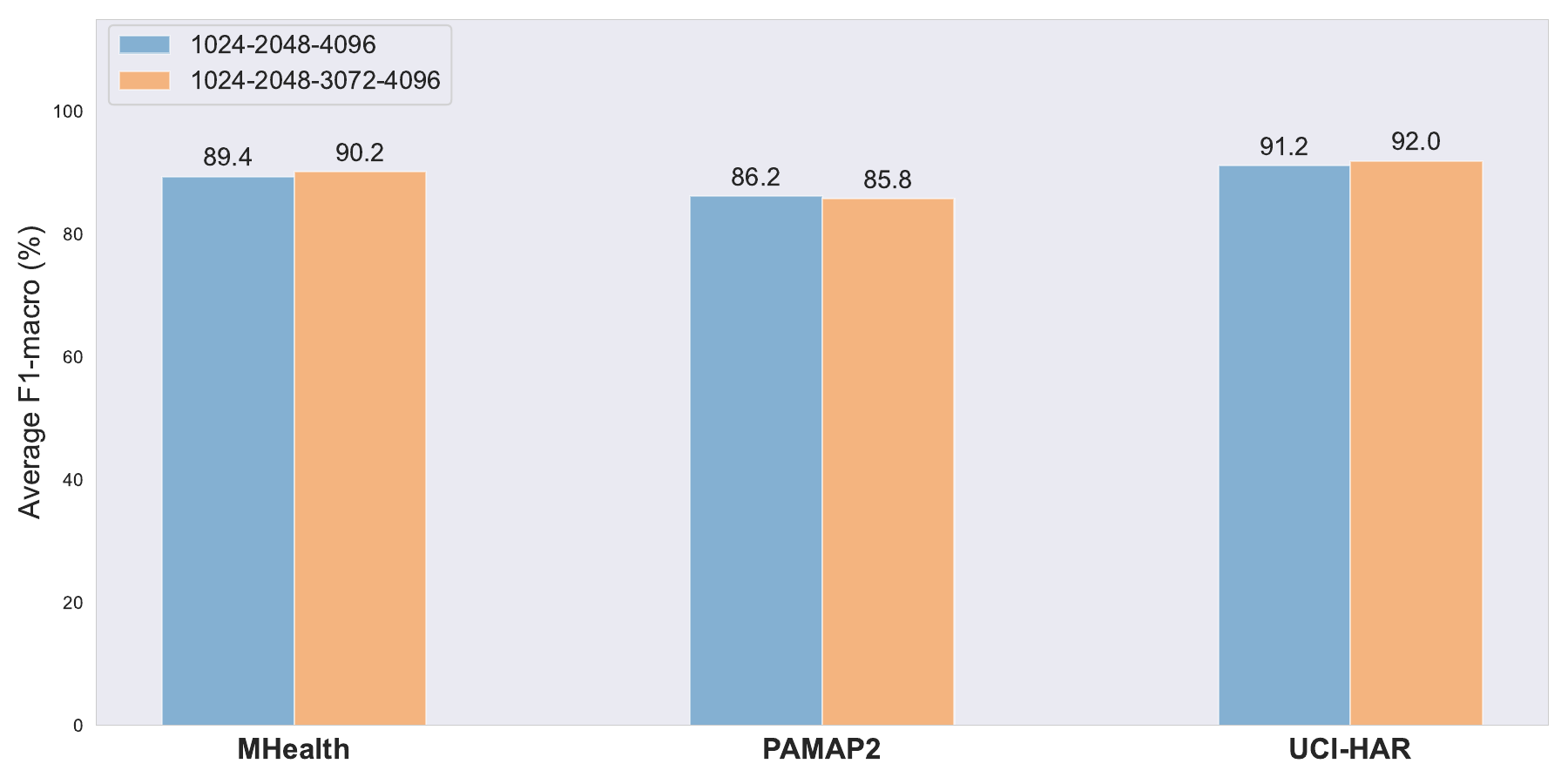}
    \caption{Effect of the number of alignment module layers.}
    \label{fig:hidden_layer_ablation}
\end{figure}

\paragraph{MLP Depth Trade-offs.}  
We examine how the depth of the alignment module MLP affects performance on UCI-HAR, PAMAP2, and MHealth. As shown in Figure~\ref{fig:hidden_layer_ablation}, increasing the number of hidden layers from one (\(1024 \to 2048 \to 4096\)) to two (\(1024 \to 2048 \to 3072 \to 4096\)) yields mixed results. F1-macro scores improve on UCI-HAR and MHealth, but slightly decrease on PAMAP2. These findings suggest that deeper MLPs do not always improve performance, and a single hidden layer offers a good balance between accuracy and efficiency.

\begin{figure}[t]
    \centering
    \includegraphics[width=\linewidth]{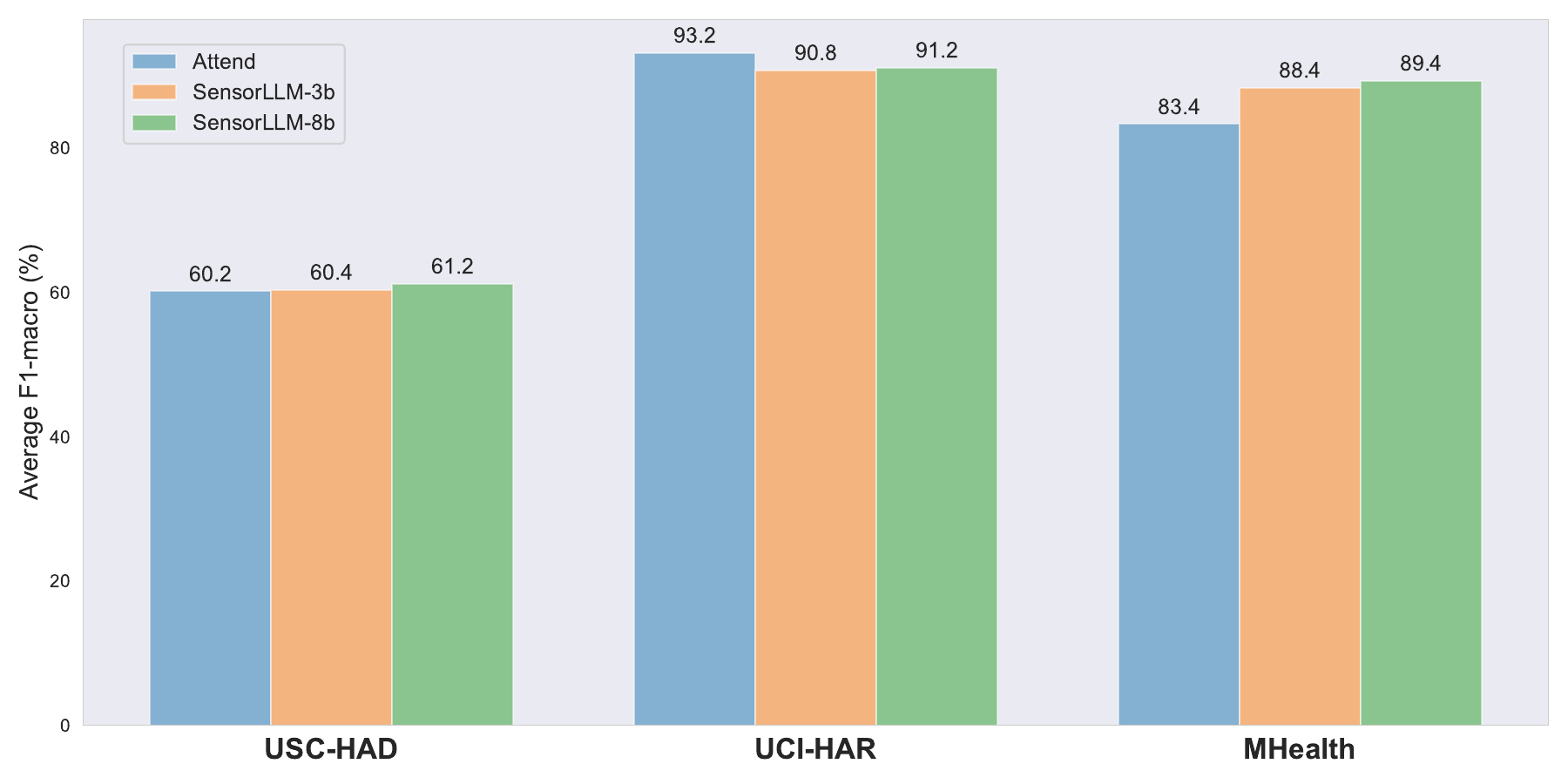}
    \caption{Effect of Model Size.}
    \label{fig:size_ablation}
\end{figure}

\paragraph{Smaller SensorLLM Still Compete.}  
To address computational feasibility for deployment in resource-constrained environments, we evaluate SensorLLM-3b, a lighter variant built with Chronos-base and LLaMA3.2-3b. Experiments were conducted on USC-HAD, UCI-HAR, and MHealth. As shown in Figure~\ref{fig:size_ablation}, SensorLLM-3b achieves slightly lower performance than SensorLLM-8b, reflecting the trade-off between model size and accuracy. Nevertheless, it remains competitive, outperforming Attend on USC-HAD and MHealth, and closely trailing it on UCI-HAR. These results suggest that SensorLLM-3b provides a strong balance between efficiency and performance, making it a viable choice for real-world, resource-limited applications.

\begin{table}[t]
    \centering
    \resizebox{\linewidth}{!}{
    \begin{tabular}{l|ccc}
        \toprule
        \multirow{3}{*}{\textbf{Dataset}}  & \multicolumn{2}{c}{\textbf{F1-macro}} & \multirow{3}{*}{\textbf{\# Channels}} \\
        \cmidrule(lr){2-3}
         & \textbf{w/o ST} & \textbf{w/ ST} & \\
        \midrule
        MHealth  & 89.6\textsubscript{\(\pm 2.70\)} & \textbf{90.2}\textsubscript{\(\pm 3.11\)} & 15 \\
        PAMAP2   & 84.4\textsubscript{\(\pm 1.14\)} & \textbf{85.8}\textsubscript{\(\pm 0.84\)} & 27 \\
        \bottomrule
    \end{tabular}
    }
    \caption{Effect of special tokens on HAR based on two-layer alignment MLP. \textit{ST} refers to special tokens.}
    \label{tab:special_token_ablation}
\end{table}

\paragraph{Special Tokens Improve Performance.}
We investigate the role of special tokens in helping SensorLLM distinguish sensor data from text and identify different sensor channel types. Special tokens are added to the aligned embeddings of each sensor channel and act as learned identifiers. They provide structural cues that help the LLM model channel-wise dependencies and reduce modality confusion. We conduct experiments on PAMAP2 and MHealth, both of which contain multiple sensor channels. As shown in Table~\ref{tab:special_token_ablation}, removing special tokens leads to a slight drop in F1-macro scores, with the performance gap tending to widen as the number of sensor channels increases. This confirms their value in preserving positional and channel-level structure within a flat token sequence.

\paragraph{Alignment Enables Generalization.}  
To assess the robustness of SensorLLM, we conduct cross-dataset experiments by training the Sensor–Language Alignment Stage on USC-HAD and the Task-Aware Tuning Stage on UCI-HAR, and vice versa. While these datasets share the same sensor channels, they differ in sensor wearing position, sampling rates and activity distributions. As shown in Table~\ref{tab:crossdata_results}, SensorLLM achieves performance comparable to models trained entirely on the same dataset. This suggests that once modality alignment is learned, it can be transferred across datasets without retraining. These results indicate that SensorLLM does not overfit to dataset-specific patterns but learns generalizable sensor-language representations, demonstrating strong cross-dataset adaptability and paving the way for more universal TS–LLM frameworks.

\begin{table}[t]
\centering
\begin{tabular}{llc}
    \toprule
    \bf Stage 2 & \bf Stage 1 & \bf F1-macro \\
    \midrule
    \multirow{2}{*}{USC-HAD} 
       & UCI-HAR  & \textbf{61.6}\textsubscript{\(\pm 2.07\)} \\
        & USC-HAD  & 61.2\textsubscript{\(\pm 3.56\)} \\  
    \midrule
    \multirow{2}{*}{UCI-HAR} 
        & UCI-HAR & \textbf{91.2}\textsubscript{\(\pm 1.48\)} \\ 
        & USC-HAD &  91.0\textsubscript{\(\pm 1.41\)} \\ 
    \bottomrule
\end{tabular}
\caption{F1-macro scores for cross-dataset experiments.}
\label{tab:crossdata_results}
\end{table}
\section{Conclusions}
\label{conclusion}
We present SensorLLM, the first multimodal framework that aligns sensor data with automatically generated text at a human-perception level, moving beyond machine-level alignment. SensorLLM effectively captures complex sensor patterns, achieving superior performance in HAR tasks. Experiments across diverse datasets demonstrate its robustness in handling variable-length sequences, multi-channel inputs, and textual metadata. Cross-dataset results further highlight its strong generalizability without requiring dataset-specific alignment. This work establishes a foundation for Sensor-Text MLLMs, with potential applications for sensor data reasoning. We release our code and data generation pipeline to facilitate future research on integrating time-series and text, particularly in low-resource domains.

\section{Limitations}
\label{limitations}
While SensorLLM demonstrates strong performance in aligning sensor data with LLMs, certain limitations remain, offering directions for future exploration.

\paragraph{Choice of Classifier.}
It is worth noting that the primary contribution of our work lies in proposing a novel TS-text alignment strategy and rigorously validating its effectiveness. To ensure fair comparisons with existing HAR baselines, we adopt a classifier for downstream tasks rather than fully exploiting the LLM’s generative abilities. While this design allows rigorous validation of our proposed alignment method, it also imposes limitations: relying on a fixed-class classifier may constrain adaptability to new activity categories and does not fully leverage the reasoning potential of LLMs. Although our framework is compatible with other downstream heads, such as a language modeling (LM) head for prompt-based generation, we leave this exploration to future work. Investigating generative or zero-shot approaches could enable broader applications, including activity discovery and open-set recognition.

\paragraph{Scope of Sensor-Text Alignment.}
Our alignment focuses on mapping sensor data to trend-descriptive text, demonstrating clear benefits for LLM-based HAR. However, human-intuitive descriptions of sensor data extend beyond trend changes, incorporating frequency-domain features, periodicity, and higher-order patterns may further enhance an LLM’s ability to interpret time-series data. Future research could investigate whether aligning text with alternative sensor characteristics improves time-series reasoning. This could expand the potential of multimodal LLM applications in sensor-driven tasks beyond activity recognition.
\section{Acknowledgements}
\label{acknowledgements}

This research includes computations using the computational cluster Wolfpack supported by School of Computer Science and Engineering at UNSW Sydney.

\bibliography{custom}
\clearpage
\pagebreak

\appendix

\section{Appendix}
\label{appendix}

\subsection{More related work}
\label{apd:related_work}

\paragraph{Deep learning in human activity recognition.}
Over the last decade, HAR has transitioned from hand-crafted feature extraction to deep learning models capable of automatic feature learning. Early work by \citet{10.1145/1964897.1964918} utilized machine learning techniques, such as decision trees and MLPs, to classify activities using features extracted from wearable sensor data. Later, \citet{10.1145/3341163.3347727} demonstrated that optimized feature extraction within the Activity Recognition Chain (ARC) could rival or outperform end-to-end deep learning models. Deep learning models, particularly CNNs and LSTMs, have since become dominant in HAR. \citet{Bevilacqua_2019} developed a CNN-based model for HAR, while \citet{7727224} introduced CNN-pf and CNN-pff architectures that apply partial and full weight sharing for better feature extraction. Other notable works include Perception-Net \citet{10.1007/978-3-030-01054-6_7}, which leverages 2D convolutions for multimodal sensor data, and InnoHAR~\citep{8598871}, which combines Inception CNN and GRUs for multiscale temporal feature learning. A dual-stream network utilizing convolutional layers and LSTM units, known as ConvLSTM, was employed by \citet{10.1145/3267305.3267533} to analyze complex temporal hierarchies with streams handling different time lengths. The combination of attention mechanisms with recurrent networks to enhance the computation of weights for hidden state outputs has also been demonstrated by DeepConvLSTM \cite{10.1007/978-3-030-01054-6_7} in capturing spatial-temporal features.

\paragraph{Large Language Models for Time-Series Forecasting.} LLMs have achieved remarkable success in text-related tasks, and their utility has expanded into time-series forecasting. \citet{xue2023promptcast} presents PromptCast, which redefines time-series forecasting as a natural language generation task by transforming numerical inputs into textual prompts, enabling pre-trained language models to handle forecasting tasks with superior generalization in zero-shot settings. \citet{gruver2024llmtime} explores encoding time-series as numerical strings, allowing LLMs like GPT-3 and LLaMA-2 to perform zero-shot forecasting, matching or surpassing the performance of specialized models, while highlighting challenges in uncertainty calibration due to model modifications like RLHF. \citet{zhou2023fits} demonstrates that pre-trained language and image models, such as a Frozen Pretrained Transformer (FPT), can be adapted for diverse time-series tasks like classification, forecasting, and anomaly detection, leveraging self-attention mechanisms to bridge the gap between different data types and achieving state-of-the-art performance across various tasks. \citet{jin2024position} highlights the transformative potential of LLMs for time-series analysis by integrating language models with traditional analytical methods. \citet{jin2023time} introduces a reprogramming framework that aligns time-series data with natural language processing capabilities, enabling LLMs to perform time-series forecasting without altering the core model structure. \citet{cao2024tempo} presents TEMPO, a generative transformer framework based on prompt tuning, which adapts pre-trained models for time-series forecasting by decomposing trends, seasonality, and residual information. \citet{sun2024test} proposes TEST, an innovative embedding technique that integrates time-series data with LLMs through instance-wise, feature-wise, and text-prototype-aligned contrast, yielding improved or comparable results across various applications. \citet{chang2024llm4ts} develops a framework that enhances pre-trained LLMs for multivariate time-series forecasting through a two-stage fine-tuning process and a novel multi-scale temporal aggregation method, outperforming traditional models in both full-shot and few-shot scenarios. Finally, \citet{liu2024unitime} introduces UniTime, a unified model that leverages language instructions and a Language-TS Transformer to handle multivariate time series across different domains, demonstrating enhanced forecasting performance and zero-shot transferability.

\subsection{Sensor–Text Data Pair Generation}
\label{apd:data_generation}

We generate text data from sensor readings using predefined sentence templates (Tables~\ref{tab:answer_templates1}, \ref{tab:q_tpl}, \ref{tab:answer_templates2}). These templates are randomly selected to create diverse question-answer (QA) pairs. To enhance variability, we employ GPT-4o to generate synonymous variations. Each sentence contains placeholders for numerical values (e.g., timestamps, sensor readings) or textual information, which are dynamically replaced to produce coherent QA pairs aligned with the sensor data.

\begin{table}[t]
\centering
\begin{tabular}{p{0.92\linewidth}}
\toprule
\textbf{Trend Description Templates} \\
\begin{itemize}[leftmargin=*]
    \item \{start\_time\}s to \{end\_time\}s: \{trend\}
    \item \{start\_time\} seconds to \{end\_time\} seconds: \{trend\}
    \item \{start\_time\} to \{end\_time\} seconds: \{trend\}
    \item \{start\_time\}-\{end\_time\} seconds: \{trend\}
    \item \{start\_time\}-\{end\_time\}s: \{trend\}
    \item \{start\_time\}s-\{end\_time\}s: \{trend\}
\end{itemize}\\
\bottomrule
\end{tabular}
\caption{Examples of answer templates used for trend descriptions.}
\label{tab:answer_templates1}
\end{table}

\begin{table}[t]
\centering
\resizebox{\linewidth}{!}{
\begin{tabular}{l|cc|cc}
\toprule
\textbf{Dataset} & \multicolumn{2}{c|}{\textbf{Stage 1}} & \multicolumn{2}{c}{\textbf{Stage 2}} \\
                 & \textbf{Train} & \textbf{Test}             & \textbf{Train} & \textbf{Test} \\
\midrule
USC-HAD        &     300,744    &    58,704        & 22,790          &      4,555      \\
UCI-HAR        &   128,292       &      25,932                &   7,352        & 2,947           \\
PAMAP2        &  738,666         &      271,674                &   14,163          &    5,210        \\
MHealth         & 283,020          & 60,780     & 4771             & 2,039           \\
CAPTURE-24        & 72,714          & 35,688     & 61,327             & 30,138           \\
\bottomrule
\end{tabular}}
\caption{Training and testing sensor-text QA pairs counts for Stage 1 and Stage 2.}
\label{tab:stage-splits}
\end{table}

\begin{table*}[htbp]
\centering
\begin{tabular}{p{0.92\linewidth}}
\toprule
\textbf{Trend Description Templates} \\
\begin{itemize}[leftmargin=*]
    \item Kindly provide a detailed analysis of the trend changes observed in the \{data\}.
    \item Please offer a comprehensive description of how the trends in the \{data\} have evolved.
    \item I would appreciate a thorough explanation of the trend fluctuations that occurred within the \{data\}.
    \item Could you examine the \{data\} in depth and explain the trend shifts observed step by step?
    \item Detail the \{data\}'s trend transitions.
    \item Could you assess the \{data\} and describe the trend transformations step by step?
    \item Could you analyze the trends observed in the \{data\} over the specified period step by step?
    \item Can you dissect the \{data\} and explain the trend changes in a detailed manner?
    \item What trend changes can be seen in the \{data\}?
\end{itemize} \\
\midrule
\textbf{Summary Templates} \\
\begin{itemize}[leftmargin=*]
    \item Could you provide a summary of the main features of the input \{data\} and the distribution of the trends?
    \item Please give an overview of the essential attributes of the input \{data\} and the spread of the trends.
    \item Describe the salient features and trend distribution within the \{data\}.
    \item Give a summary of the \{data\}'s main elements and trend apportionment.
    \item Summarize the \{data\}'s core features and trend dissemination.
    \item Outline the principal aspects and trend allocation of the \{data\}.
    \item Summarize the key features and trend distribution of the \{data\}.
    \item I need a summary of \{data\}'s main elements and their trend distributions.
\end{itemize} \\
\bottomrule
\end{tabular}
\caption{Examples of question templates used for trend description and summary generation.}
\label{tab:q_tpl}
\end{table*}

\begin{table*}[htbp]
\centering
\begin{tabular}{p{0.92\linewidth}}
\toprule
\textbf{Summary 1: Trend Count} \\
\begin{itemize}[leftmargin=*]
    \item Number of \{trend\} trends: \{num\}
    \item Count of \{trend\} trends: \{num\}
    \item Number of \{trend\} segments: \{num\}
    \item Count of \{trend\} segments: \{num\}
\end{itemize} \\
\midrule
\textbf{Summary 2: Sensor Data Context} \\
\begin{itemize}[leftmargin=*]
    \item The given \{data\_name\} represents \{sensor\_name\} sensor readings from \{start\_time\}s to \{end\_time\}s.
    \item The \{data\_name\} contains \{sensor\_name\} sensor readings recorded between \{start\_time\} and \{end\_time\} seconds.
    \item The \{sensor\_name\} sensor readings collected from \{start\_time\} to \{end\_time\} seconds are presented in this \{data\_name\}.
\end{itemize} \\
\midrule
\textbf{Summary 3: Trend Change Statistics} \\
\begin{itemize}[leftmargin=*]
    \item The data exhibits \{trend\_num\} distinct trends, with \{change\_num\} trend changes observed.
    \item Across \{trend\_num\} trends, the data shows \{change\_num\} occurrences of trend shifts.
    \item \{trend\_num\} trends are present, with \{change\_num\} instances of trend changes.
\end{itemize} \\
\midrule
\textbf{Summary 4: Cumulative Trend Analysis} \\
\begin{itemize}[leftmargin=*]
    \item To sum up, the data exhibited a \{trend\_type\} trend for a total duration of \{total\_time\} seconds.
    \item Overall, the data showed a \{trend\_type\} trend spanning \{total\_time\} seconds.
    \item In conclusion, the trend was \{trend\_type\} over \{total\_time\} seconds.
\end{itemize} \\
\midrule
\textbf{Summary 5: Overall Trend Summary} \\
\begin{itemize}[leftmargin=*]
    \item The overall trend is \{overall\_trend\}.
    \item The primary trend detected is \{overall\_trend\}.
    \item Looking at the broader pattern, the trend is \{overall\_trend\}.
\end{itemize} \\
\bottomrule
\end{tabular}
\caption{Examples of answer templates used for summaries.}
\label{tab:answer_templates2}
\end{table*}

The system prompt instructs the model on how to respond to generated questions, incorporating dataset-specific attributes such as sensor frequency and sampling rate. These tailored prompts ensure responses align with the unique characteristics of each dataset. Below is the system prompt template used for all datasets:
\begin{itemize}
    \item A dialogue between a researcher and an AI assistant. The AI analyzes a sensor time-series dataset (\textit{N} points, sampled at \{sample\_rate\}Hz) to answer specific questions, demonstrating its analytical capabilities and the potential for human-AI collaboration in interpreting sensor data.
\end{itemize}

\paragraph{Sensor-Language Alignment Stage}focuses on aligning uni-variate sensor sequence of variable length with descriptive textual responses and includes two types of QA tasks:
\begin{itemize}
    \item \textbf{Trend Analysis QA}, which describes how the signal changes within the window.
    \item \textbf{Trend Summary QA}, which summarizes the overall behavior across a window in a concise natural language phrase.
\end{itemize}

\paragraph{Task-Aware Tuning Stage}focuses on using multi-variate sensor sequences to perform human activity classification, leveraging the aligned representations learned in the alignment stage. This stage contains statistical information from each sensor channel as part of the input representation.

The distribution of training and testing data across both stages is summarized in Table~\ref{tab:stage-splits}.

\subsection{Chronos}
\label{apd:chronos}

Chronos~\citep{ansari2024chronos} is a pretrained probabilistic time-series framework that tokenizes real-valued time-series data into discrete representations for language model training. It utilizes scaling and quantization to transform time-series data into a fixed vocabulary, enabling T5-based~\citep{2020t5} models to learn from tokenized sequences using cross-entropy loss. Pretrained on diverse public and synthetic datasets, Chronos surpasses existing models on familiar datasets and demonstrates strong zero-shot performance on unseen tasks, making it a versatile tool for time-series forecasting across domains.

\paragraph{Time-Series Tokenization and Quantization.} 
Chronos converts time-series data into discrete tokens through a two-step process: normalization and quantization. Mean scaling is first applied to ensure consistency across different time series:

\begin{equation}
\tilde{x} = \frac{x}{\text{mean}(\lvert x \rvert)}
\end{equation}

Next, the normalized values are quantized using $B$ bin centers \( c_1, \dots, c_B \) and corresponding bin edges \( b_1, \dots, b_{B-1} \), mapping real values to discrete tokens via:

\begin{equation}
q(x) =
\begin{cases}
1 & \text{if} \ -\infty \leq x < b_1, \\
2 & \text{if} \ b_1 \leq x < b_2, \\
\vdots \\
B & \text{if} \ b_{B-1} \leq x < \infty.
\end{cases}
\end{equation}

Special tokens such as \texttt{PAD} and \texttt{EOS} are added to handle sequence padding and denote the end of sequences, allowing Chronos to process variable-length inputs efficiently within language models.

\paragraph{Objective Function.} 
Chronos models the tokenized time series using a categorical distribution over the vocabulary \( V_{ts} \), minimizing the cross-entropy loss:

\begin{equation}
\begin{split}
\ell(\theta) = - \sum_{h=1}^{H+1} \sum_{i=1}^{|V_{ts}|} 
\mathbf{1} (z_{C+h+1} = i) \\
\cdot \log p_\theta(z_{C+h+1} = i \mid z_{1:C+h})
\end{split}
\end{equation}

where \( C \) is the historical context length, \( H \) is the forecast horizon, and \( p_\theta \) is the predicted token distribution.

This approach offers two key advantages: (i) Seamless integration with language models, requiring no architectural modifications, and (ii) Flexible distribution learning, enabling robust generalization across diverse time-series datasets.

\subsection{GPT-4o Prompt for Sensor Data Trend Analysis}
\label{apd:gpt_gen}

Table~\ref{apd:gpt_gen_table} presents the system prompt used to generate trend-descriptive texts from sensor data, providing a structured framework for GPT-4o to analyze and respond to specific questions. This standardized prompt ensures consistency in GPT-4o's interpretation of time-series data, allowing direct comparison with descriptions produced by SensorLLM.

\begin{table}[h]
\centering
\begin{tabular}{p{0.15\linewidth}p{0.7\linewidth}}
\toprule
\textbf{Prompt} & A dialogue between a curious researcher and an AI assistant. The AI analyzes a sensor time-series dataset (N points, \{sr\}Hz sampling rate) to answer specific questions. \\
& \\
& Please output your answer in the format like this example:\\
& \{example from ground-truth\} \\
& \\
& Now, analyze the following: \\
& Input: \{sensor\_data\} How trends in the given sensor data evolve? \\
& Output: \\
\bottomrule
\end{tabular}
\caption{Prompt for GPT-4o to generate descriptive texts based on the given numerical sensor data.}
\label{apd:gpt_gen_table}
\end{table}

We evaluate GPT-4o’s ability to interpret numerical sensor data by assessing its responses against human evaluations and NLP metrics. This comparison benchmarks GPT-4o’s performance against SensorLLM, highlighting differences in how both models process time-series data trends. The results demonstrate the effectiveness of SensorLLM’s Sensor-Language Alignment Stage.

\subsection{Evaluation Metrics for Sensor-Language Alignment Stage }
\label{apd:metrics}
In this section, we describe the various evaluation metrics used to assess the performance of SensorLLM in generating trend descriptions from sensor data. Each metric offers a distinct perspective on model performance, ranging from surface-level textual similarity to more complex semantic alignment.

\paragraph{BLEU-1~\citep{10.3115/1073083.1073135}.} BLEU (Bilingual Evaluation Understudy) is a precision-based metric commonly used to evaluate machine-generated text by comparing it to reference texts. BLEU-1 focuses on unigram (single-word) overlap, assessing the lexical similarity between the generated and reference text. While useful for measuring word-level matches, BLEU-1 does not capture deeper semantic meaning, making it most effective for surface-level alignment.

\paragraph{ROUGE-1 and ROUGE-L~\citep{lin-2004-rouge}.} ROUGE (Recall-Oriented Understudy for Gisting Evaluation) evaluates the recall-oriented overlap between generated text and reference text. ROUGE-1 focuses on unigram recall, similar to BLEU-1 but emphasizing how much of the reference text is captured. ROUGE-L measures the longest common subsequence, assessing both precision and recall in terms of structure and content overlap, though it does not evaluate semantic accuracy.

\paragraph{METEOR~\citep{banerjee-lavie-2005-meteor}.} METEOR (Metric for Evaluation of Translation with Explicit Ordering)combines precision and recall, with additional alignment techniques such as stemming and synonym matching. Unlike BLEU and ROUGE, METEOR accounts for some degree of semantic similarity. However, its emphasis is still on word-level alignment rather than factual accuracy or meaning.

\paragraph{SBERT~\citep{Reimers2019SentenceBERTSE}.} SBERT (Sentence-BERT)~\footnote{https://huggingface.co/sentence-transformers/all-mpnet-base-v2} is a metric that generates sentence embeddings using the BERT architecture. It computes cosine similarity between embeddings of the generated and reference texts, providing a deeper assessment of semantic similarity beyond lexical matches.

\paragraph{SimCSE~\citep{gao2021simcse}.} SimCSE (Simple Contrastive Sentence Embedding)~\footnote{https://huggingface.co/princeton-nlp/sup-simcse-roberta-large} introduces a contrastive learning approach to fine-tune language models for sentence embeddings. By applying different dropout masks to the same sentence, it generates positive examples, encouraging similar embeddings for semantically identical sentences while distinguishing different ones.

\paragraph{GPT-4o Evaluation.} In addition to the NLP metrics, we also employed GPT-4o as a human-like evaluator. Given its strong reasoning and comprehension abilities, GPT-4o was tasked with scoring the generated text based on its alignment with the ground truth. GPT-4o evaluated the correctness, completeness, and coherence of the trend descriptions and assigned a score from 1 to 5, accompanied by an explanation (see Table~\ref{apd:eval_gpt}). This type of evaluation provides insights into how well the generated outputs capture the nuances of sensor data trends in a manner similar to human understanding.

\begin{table*}[htbp]
\centering
\begin{tabular}{p{0.2\linewidth}p{0.7\linewidth}}
\toprule
\textbf{Prompt} & Please evaluate the model-generated trend descriptions against the ground truth. Rate each pair based on the degree of accuracy, using a scale from 1 to 5, where 1 represents the lowest correctness and 5 represents the highest. Deduct 1 point for minor errors in the trend description, and 2-3 points for moderate errors. \\
& \\
& Provide your score (1-5) and a brief explanation in the format: "score\#reason" (e.g., 4\#The description of trend changes slightly differs from the ground truth).\\
& \\
& Now, please proceed to score the following: \\
& Model: \{model\_output\} \\
& Human: \{ground\_truth\} \\
& Output: \\
\midrule
\textbf{Output example 1}: & 2\#Significant discrepancies in segment durations and trend counts compared to ground-truth. \\
\midrule
\textbf{Output example 2}: & 5\#The model's description matches the human-generated text accurately. \\
\bottomrule
\end{tabular}
\caption{Prompt and output examples for GPT-4o in evaluating model-generated texts and ground-truth.}
\label{apd:eval_gpt}
\end{table*}

\paragraph{Human Evaluation.} Finally, five human experts assessed the correctness and quality of the generated trend descriptions. Following the same criteria as GPT-4o, they rated the outputs on a scale from 1 to 5, focusing on the factual accuracy and coherence of the descriptions. This manual evaluation serves as an important benchmark for the model's performance from a human perspective, ensuring that the generated outputs are not only technically correct but also practically useful for human interpretation.

\subsection{Datasets}
\label{apd:dataset}
We used five datasets in our study:
\paragraph{USC Human Activity Dataset (USC-HAD).}
USC-HAD~\citep{10.1145/2370216.2370438} consists of six sensor readings from body-worn 3-axis accelerometers and gyroscopes, collected from 14 subjects. The data is sampled at 100 Hz across six channels and includes 12 activity class labels. For evaluation, we use data from subjects 13 and 14 as the test set, while the remaining subjects' data are used for training. A window size $w \in [5, 200]$ is used in alignment stage, and $w = 200$ with stride of $100$ are used in HAR.

\paragraph{UCI Human Activity Recognition Dataset (UCI-HAR).}
UCI-HAR~\citep{Anguita2013APD} includes data collected from 30 volunteers performing six activities while wearing a smartphone on their waist. The embedded accelerometer and gyroscope sensors sampled data at 50 Hz across six channels. The dataset was partitioned into 70\% for training and 30\% for testing. A window size $w \in [5, 200]$ is used in alignment stage, and $w = 128$ with stride of $64$ is used in HAR.

\paragraph{Physical Activity Monitoring Dataset (PAMAP2).}
 PAMAP2~\citep{6246152} includes data from nine subjects wearing IMUs on their chest, hands, and ankles. IMUs capture the acceleration, gyroscope, and magnetometer data across 27 channels and include 12 activity class labels. For our experiments, data from subjects 105 and 106 are used as the test set, with the remaining subjects' data used for training. The sample rate is downsampled from 100 Hz to 50 Hz. A window size $w \in [5, 100]$ is used in alignment stage, and $w = 100$ with stride of $50$ in HAR.

\paragraph{Mobile Health Dataset (MHealth).}
MHealth~\citep{Baos2014mHealthDroidAN} contains body motion and vital sign recordings from ten volunteers. Sensors were placed on the chest, right wrist, and left ankle of each subject. For our experiments, we used acceleration data from the chest, left ankle, and right lower arm, along with gyroscope data from the left ankle and right lower arm, resulting in a total of 15 channels. The data is sampled at 50 Hz and includes 12 activity class labels. Data from subjects 1, 3, and 6 is used as the test set, while the remaining subjects' data are used for training. We use a window size $w \in [5, 100]$ in alignment stage and $w = 100$ with stride of $50$ in HAR.

\paragraph{CAPTURE-24.}
CAPTURE-24~\citep{chan2024capture24largedatasetwristworn} is a large-scale dataset featuring 3-channel wrist-worn accelerometer data collected in free-living settings for over 24 hours per participant. It includes annotated data from 151 participants, making it significantly larger than existing datasets. We used the first 100 participants as the training set and the remaining 51 as the test set. For each subject, sequences were windowed, and 5\% of the data was randomly selected for training and testing. The sample rate was downsampled from 100 Hz to 50 Hz and it includes 10 activity class labels. During the alignment stage, we used a variable window size $w \in [10, 500]$, while in the HAR, we fixed $w = 500$ with a stride of $250$.

Each dataset includes multiple activity classes, and the proportion of each class in the dataset is shown in Table~\ref{tab:dataset_categories}.

\begin{table*}[htbp]
\centering
\begin{tabular}{|c|c|l|l|}
\hline
\textbf{Dataset} & \textbf{\# Classes} & \textbf{Classes} & \textbf{Proportions (\%)} \\ \hline
USC-HAD & 12 & 
\begin{tabular}[c]{@{}l@{}}Sleeping, Sitting, Elevator down, \\ Elevator up, Standing, Jumping, \\ Walking downstairs, Walking right, \\ Walking forward, Running forward, \\ Walking upstairs, Walking left\end{tabular} & 
\begin{tabular}[c]{@{}l@{}}12.97, 9.06, 6.04, 5.94, 8.6, 3.62, \\ 7.61, 9.81, 13.15, 5.72, 8.22, 9.25\end{tabular} \\ \hline

UCI-HAR & 6 & 
\begin{tabular}[c]{@{}l@{}}Standing, Sitting, Laying, \\ Walking, Walking downstairs, \\ Walking upstairs\end{tabular} & 
\begin{tabular}[c]{@{}l@{}}18.69, 17.49, 19.14, \\ 16.68, 13.41, 14.59\end{tabular} \\ \hline

PAMAP2 & 12 & 
\begin{tabular}[c]{@{}l@{}}Lying, Sitting, Standing, \\ Ironing, Vacuum cleaning, \\ Ascending stairs, Descending stairs, \\ Walking, Nordic walking, Cycling, \\ Running, Rope jumping\end{tabular} & 
\begin{tabular}[c]{@{}l@{}}10.25, 9.52, 10.11, \\ 11.82, 9.14, 6.3, \\ 5.67, 12.77, 9.52, \\ 8.42, 3.57, 2.91\end{tabular} \\ \hline

MHealth & 12 & 
\begin{tabular}[c]{@{}l@{}}Climbing stairs, Standing still, \\ Sitting and relaxing, Lying down, \\ Walking, Waist bends forward, \\ Frontal elevation of arms, \\ Knees bending (crouching), \\ Jogging, Running, Jump front \\ \& back, Cycling\end{tabular} & 
\begin{tabular}[c]{@{}l@{}}8.91, 8.95, 8.95, \\ 8.95, 8.95, 8.26, \\ 8.7, 8.53, 8.95, \\ 8.95, 2.96, 8.95\end{tabular} \\ \hline

CAPTURE-24 & 10 & 
\begin{tabular}[c]{@{}l@{}}Sleep, Household-chores, Walking, \\ Vehicle, Standing, Mixed-activity, \\ Sitting, Bicycling, Sports, \\ Manual-work\end{tabular} & 
\begin{tabular}[c]{@{}l@{}}37.45, 6.5, 6.16, \\ 3.83, 3.25, 3.49, \\ 37.07, 1.03, 0.43, 0.79\end{tabular} \\ \hline
\end{tabular}
\caption{Dataset classes and Proportions}
\label{tab:dataset_categories}
\end{table*}

\subsection{Baselines for Task-Aware Tuning Stage}
\label{apd:baselines}

In Task-Aware Tuning Stage, we compare SensorLLM against several state-of-the-art baseline models for time-series classification and human activity recognition (HAR). These models were selected for their strong performance in relevant tasks, providing a thorough benchmark for evaluating SensorLLM’s effectiveness.

\paragraph{Transformer~\citep{NIPS2017_3f5ee243}.} The Transformer model is a widely-used architecture in various tasks, including time-series forecasting and classification. It uses self-attention mechanisms to capture long-range dependencies in sequential data, making it highly effective for modeling complex temporal relationships.

\paragraph{Informer~\citep{haoyietal-informer-2021}.} Informer is a transformer-based model designed for long sequence time-series data. It addresses key limitations of standard Transformers, such as high time complexity and memory usage, through three innovations: ProbSparse self-attention, which reduces time complexity; self-attention distilling, which enhances efficiency by focusing on dominant patterns; and a generative decoder that predicts entire sequences in a single forward pass. 

\paragraph{NS-Transformer~\citep{liu2022non}.} Non-stationary Transformers (NS-Transformer) tackles the issue of over-stationarization in time-series by balancing series predictability and model capability. It introduces Series Stationarization to normalize inputs and De-stationary Attention to restore intrinsic non-stationary information into temporal dependencies. 

\paragraph{PatchTST~\citep{Yuqietal-2023-PatchTST}.} PatchTST is a Transformer-based model for multivariate time series tasks, using subseries-level patches as input tokens and a channel-independent approach to reduce computation and improve efficiency. This design retains local semantics and allows for longer historical context, significantly improving long-term forecasting accuracy. 

\paragraph{TimesNet~\citep{wu2023timesnet}.} TimesNet is a versatile backbone for time series analysis that transforms 1D time series into 2D tensors to better capture intraperiod and interperiod variations. This 2D transformation allows for more efficient modeling using 2D kernels. It also introduces TimesBlock to adaptively discovers multi-periodicity and extracts temporal features from transformed 2D tensors using a parameter-efficient inception block.

\paragraph{iTransformer~\citep{liu2024itransformer}.} iTransformer reimagines the Transformer architecture by applying attention and feed-forward networks to inverted dimensions. Time points of individual series are embedded as variate tokens, allowing the attention mechanism to capture multivariate correlations, while the feed-forward network learns nonlinear representations for each token.

\paragraph{DeepConvLSTM~\citep{s16010115}.} DeepConvLSTM integrates four consecutive convolutional layers followed by two LSTM layers to effectively capture both spatial and temporal dynamics in sensor data. The final output vector is passed through a fully connected layer, and the softmax function is applied to produce activity class probabilities as the model's final output.

\paragraph{DeepConvLSTMAttn~\citep{10.1145/3267242.3267287}.} DeepConvLSTMAttn enhances the original DeepConvLSTM by integrating an attention mechanism to improve temporal modeling in HAR tasks. Instead of using the last LSTM hidden state for classification, the attention mechanism is applied to the first 7 hidden states, representing historical temporal context. These states are transformed through linear layers to generate attention scores, which are passed through softmax to produce weights. The weighted sum of the hidden states is combined with the last hidden state to form the final embedding for classification.

\paragraph{Attend~\citep{10.1145/3448083}.} The Attend model use the latent relationships between multi-channel sensor modalities and specific activities, apply data-agnostic augmentation to regularize sensor data streams, and incorporate a classification loss criterion to minimize intra-class representation differences while maximizing inter-class separability. These innovations result in more discriminative activity representations, significantly improving HAR performance. 

\paragraph{Chronos+MLP.} Chronos~\citep{ansari2024chronos}+MLP is a baseline designed to evaluate whether the performance gains in SensorLLM are solely attributable to Chronos and the MLP. In SensorLLM, Chronos is used to generate sensor embeddings, which are then mapped by the MLP for input into the LLM to perform HAR. Since Chronos does not natively support classification tasks and only processes single-channel data, we adapt it for HAR by inputting each channel's data separately into Chronos. The resulting sensor embeddings for all channels are then concatenated and fed into an MLP, which acts as a classifier. This setup allows us to benchmark against a simpler framework and validate the unique contributions of SensorLLM's design.

\paragraph{GPT4TS~\citep{zhou2023fits}.}
GPT4TS is a unified framework that leverages a frozen pre-trained language model (e.g., GPT-2~\citep{radford2019language}) to achieve state-of-the-art or comparable performance across various time-series analysis tasks, including classification, forecasting (short/long-term), imputation, anomaly detection, and few-shot/zero-sample forecasting. The authors also found that self-attention functions similarly to PCA, providing a theoretical explanation for the versatility of transformers.

\subsection{Evaluation Metrics for Task-Aware Tuning Stage}
\label{apd:metrics_stage2}
In our evaluation, we use the F1-macro score to assess the model's performance across datasets. F1-macro is particularly suitable for datasets with imbalanced label distributions, which is common in Human Activity Recognition (HAR) tasks where certain activities are overrepresented while others have fewer samples. Unlike the micro F1 score, which emphasizes the performance on frequent classes, F1-macro treats each class equally by calculating the F1 score independently for each class and then averaging them.

The formula for the F1-macro score is:
\begin{equation}
\text{F1-macro} = \frac{1}{C} \sum_{i=1}^{C} \text{F1}_i
\end{equation}

\noindent where \(C\) is the total number of classes, and \(\text{F1}_i\) is the F1 score for class \(i\). The F1 score for each class is calculated as:

\begin{equation}
\text{F1}_i = \frac{2 \times \text{Precision}_i \times \text{Recall}_i}{\text{Precision}_i + \text{Recall}_i}
\end{equation}

\noindent The precision and recall for each class are defined as:

\begin{equation}
\text{Precision}_i = \frac{\text{TP}_i}{\text{TP}_i + \text{FP}_i}
\end{equation}

\begin{equation}
\text{Recall}_i = \frac{\text{TP}_i}{\text{TP}_i + \text{FN}_i}
\end{equation}

\noindent where \(\text{TP}_i\), \(\text{FP}_i\), and \(\text{FN}_i\) represent the number of true positives, false positives, and false negatives for class \(i\), respectively. This metric ensures that performance is evaluated fairly across all classes, regardless of the frequency of each label, making it a robust measure for imbalanced datasets.

\subsection{Sensor-Language Alignment Stage Output Examples}
\label{apd:stage1_results}
Tables~\ref{tab:stage1_ouput_1} and ~\ref{tab:stage1_ouput_2} present two examples of the trend analysis results generated by SensorLLM and GPT-4o based on the input sensor data. From the results, it is evident that SensorLLM outperforms GPT-4o across both shorter and medium-length sequences. This demonstrates that our approach enables LLMs to better understand numerical variations, as well as accurately compute the time duration represented by the input sequences based on their length and the given sample rate. In contrast, current large language models struggle with directly interpreting numerical data, as their tokenization methods are not well-suited for tasks such as comparing numerical values or counting~\citep{yehudai2024transformerscountn}.

\onecolumn
\begin{longtable}{p{0.2\linewidth}p{0.7\linewidth}}
\toprule
\textbf{Sensor readings: } & [-9.8237,  -9.4551, -10.007,  -11.273,  -11.258,  -11.677,  -11.774, -11.638, -11.195,  -11.087,  -10.833,  -11.044,  -11.393,  -11.943,  -12.168,  -15.455, -12.967, -12.326,  -12.515,  -13.195,  -12.634,  -11.873,  -12.002,  -11.583, -10.859,  -10.349,   -9.831,   -9.1622,  -8.2721,  -6.9299, -6.255, -5.5998] \\
\midrule
\textbf{Ground-truth}: & 0.0 seconds to 0.02 seconds: growing   \\
&   0.02 seconds to 0.06 seconds: declining   \\
&   0.06 seconds to 0.08 seconds: growing   \\
&   0.08 seconds to 0.12 seconds: declining   \\
&   0.12 seconds to 0.2 seconds: growing   \\
&   0.2 seconds to 0.3 seconds: declining   \\
&   0.3 seconds to 0.34 seconds: growing   \\
&   0.34 seconds to 0.38 seconds: declining   \\
&   0.38 seconds to 0.42 seconds: growing   \\
&   0.42 seconds to 0.44 seconds: declining   \\
&   0.44 seconds to 0.62 seconds: growing   \\
&      \\
&   Total growing trends: 6   \\
&   Total declining trends: 5   \\
&      \\
&   From 0.0s to 0.62s, normalized left-ankle y-axis accelerometer data is showcased in the sensor data. Examining the data, we notice 2 clear trend characteristics, with the trend fluctuating a total of eleven times. The analysis reveals that the data's declining inclination persisted for a total of 0.24 seconds, and a growing trend within a span of 0.38 seconds. The general trend observed is growing.    \\
\midrule \midrule
\textbf{SensorLLM}: & 0.0 to 0.02 seconds: ascending \\
& 0.02 to 0.08 seconds: descending \\
& 0.08 to 0.1 seconds: ascending \\
& 0.1 to 0.12 seconds: descending \\
& 0.12 to 0.2 seconds: ascending \\
& 0.2 to 0.28 seconds: descending \\
& 0.28 to 0.32 seconds: ascending \\
& 0.32 to 0.36 seconds: descending \\
& 0.36 to 0.4 seconds: ascending \\
& 0.4 to 0.42 seconds: descending \\
& 0.42 to 0.62 seconds: ascending \\
&  \\
& Count of ascending segments: 6 \\
& Count of descending segments: 5 \\
&  \\
& The time series data encapsulates normalized left-ankle y-axis accelerometer sensor readings from 0.0 seconds to 0.62 seconds. Two separate trends and nine trend shifts are observed in the data. The analysis reveals that the data's descending inclination persisted for a total of 0.22 seconds, and an ascending trend for a sum of 0.40 seconds. The trend overall is ascending. \\
\midrule
\textbf{GPT-4o Evaluation}: &  3\#Moderate errors in trend durations and shifts compared to ground truth.   \\
\midrule \midrule
\textbf{GPT-4o's output}: &  0.0s to 0.02s: rising   \\
&  0.02s to 0.06s: falling   \\
&  0.06s to 0.08s: rising   \\
&  0.08s to 0.1s: falling   \\
&  0.1s to 0.12s: rising   \\
&  0.12s to 0.14s: falling   \\
&  0.14s to 0.3s: rising   \\
&   \\
&  Total rising segments: 4   \\
&  Total falling segments: 3   \\
&   \\
&  The normalized left-ankle y-axis accelerometer sensor readings recorded within the 0.0 to 0.3 second timeframe are presented in this sensor data. The input data displays three individual trends, with a comprehensive change count reaching 7. The examination reveals that the data's falling inclination endured for an aggregate of 0.08 seconds, succeeded by a rising trend for a cumulative period of 0.22 seconds, and a steady pattern for a total of 0.00 seconds. The dominant trend is rising. \\ 
\midrule
\textbf{GPT-4o Evaluation}: & 2\#Significant discrepancies in trend durations and counts between model and ground-truth.   \\
\bottomrule
\caption{Example 1 of SensorLLM and GPT-4o Generated Descriptions with Ground-Truth and GPT-4o Evaluation}
\label{tab:stage1_ouput_1} 
\end{longtable}

\begin{longtable}{p{0.2\linewidth}p{0.7\linewidth}}
\toprule
\textbf{Sensor readings: } & [0.53137, 0.53137, 0.53137, 0.51176, 0.51176, 0.51176, 0.45098, 0.45098, 0.45098, 0.45098, 0.45882, 0.45882, 0.45882] \\
\midrule 
\textbf{Ground-truth}: & 0.0 seconds to 0.04 seconds: stable \\
& 0.04 seconds to 0.06 seconds: decreasing \\
& 0.06 seconds to 0.1 seconds: stable \\
& 0.1 seconds to 0.12 seconds: decreasing \\
& 0.12 seconds to 0.18 seconds: stable \\
& 0.18 seconds to 0.2 seconds: increasing \\
& 0.2 seconds to 0.24 seconds: stable \\
&  \\
& Number of stable trends: 4 \\
& Number of decreasing trends: 2 \\
& Number of increasing trends: 1 \\
&  \\
& The sensor data represents readings taken from a normalized right-lower-arm x-axis gyroscope sensor between 0.0 and 0.24 seconds. Analysis reveals three separate trends within the data, undergoing a cumulative total of seven shifts in direction. Encapsulating the outcomes, the data's decreasing trend stretched across a total time of 0.04 seconds, came after an increasing pattern observed over 0.02 seconds, and a stable trend for 0.18 seconds in total. The dominant trend is decreasing.    \\
\midrule \midrule
\textbf{SensorLLM}: & 0.0s to 0.04s: consistent \\
& 0.04s to 0.06s: downward \\
& 0.06s to 0.1s: consistent \\
& 0.1s to 0.12s: downward \\
& 0.12s to 0.18s: consistent \\
& 0.18s to 0.2s: upward \\
& 0.2s to 0.24s: consistent \\
&  \\
& Number of consistent segments: 4 \\
& Number of downward segments: 2 \\
& Number of upward segments: 1 \\
 \\
& The sensor data illustrates normalized right-lower-arm x-axis gyroscope sensor readings between 0.0 and 0.24 seconds. The input data displays three individual trends, with a comprehensive change count reaching 7. To encapsulate, the data's downward trend spanned a combined duration of 0.04 seconds, and then an upward pattern for a sum of 0.02 seconds, and a consistent trend for an accumulated time of 0.18 seconds. The overarching trend is characterized as downward. \\
\midrule
\textbf{GPT-4o Evaluation}: &  5\#The model's description matches the ground truth accurately.   \\
\midrule \midrule
\textbf{GPT-4o's output}: &  0.0s to 0.06s: steady \\
&  0.06s to 0.12s: falling \\
&  0.12s to 0.14s: steady \\
&  0.14s to 0.16s: rising \\
&  0.16s to 0.18s: steady \\
&   \\
&  Total steady segments: 3 \\
&  Total rising segments: 1 \\
&  Total falling segments: 1 \\
&   \\
&  The normalized right-lower-arm x-axis gyroscope sensor readings recorded within the 0.0 to 0.18 second timeframe are presented in this sensor data. The input data displays three individual trends, with a comprehensive change count reaching 5. The examination reveals that the data's falling inclination endured for an aggregate of 0.06 seconds, succeeded by a rising trend for a cumulative period of 0.02 seconds, and a steady pattern for a total of 0.10 seconds. The dominant trend is steady. \\ 
\midrule
\textbf{GPT-4o Evaluation}: & 2\#Significant discrepancies in trend durations and counts compared to ground-truth.  \\
\bottomrule
\caption{Example 2 of SensorLLM and GPT-4o Generated Descriptions with Ground-Truth and GPT-4o Evaluation}
\label{tab:stage1_ouput_2}
\end{longtable}

\twocolumn 

\end{document}